\documentclass[11pt]{article}

\usepackage[preprint]{acl}
\usepackage{times}
\usepackage{latexsym}
\usepackage{microtype}
\usepackage{graphicx}
\usepackage{subfigure}
\usepackage{caption}
\usepackage{float}
\usepackage{subcaption}
\usepackage{xcolor} 
\usepackage{colortbl}
\usepackage{booktabs} 
\usepackage{enumitem}
\usepackage{hyperref}
\usepackage{mathtools}
\usepackage{pifont}
\usepackage{multirow}
\usepackage{amsthm}
\usepackage{amsmath}
\usepackage{amssymb}
\usepackage[textsize=tiny]{todonotes}
\usepackage{mathtools}
\usepackage[T1]{fontenc}
\usepackage{microtype}

\usepackage{inconsolata}

\usepackage[utf8]{inputenc}
\usepackage{booktabs}
\usepackage{wrapfig}

\newcommand{\cmark}{\textcolor{green!80!black}{\ding{51}}}
\newcommand{\xmark}{\textcolor{red}{\ding{55}}}

\usepackage[capitalize,noabbrev]{cleveref}
\theoremstyle{plain}

\theoremstyle{definition}

\theoremstyle{remark}

\definecolor{table-blue}{RGB}{173, 216, 230}
\definecolor{Gray}{gray}{0.95}
\definecolor{darkspringgreen}{rgb}{0.09, 0.45, 0.27}
\definecolor{americanrose}{rgb}{1.0, 0.01, 0.24}
\definecolor{darkred}{RGB}{177, 38, 26}
\definecolor{darkblue}{RGB}{67, 116, 177}
\definecolor{darkgreen}{rgb}{0.0, 0.5, 0.0}
\definecolor{lavender}{RGB}{200, 190, 230}
\definecolor{amethyst}{RGB}{180, 160, 210}
\definecolor{sakura}{RGB}{230, 200, 220}
\definecolor{myalpha}{HTML}{D62728} 
\definecolor{myentropy}{HTML}{1F77B4}
\definecolor{ForestGreen}{HTML}{228B22}
\definecolor{mycolor}{RGB}{166,200,16}

\usepackage[textsize=tiny]{todonotes}
\usepackage[most,skins,theorems]{tcolorbox}
\tcbset{
  takeawaysbox/.style={
    title=Takeaway, 
    colback=mycolor!5!white,
    colframe=mycolor!65!black,
    fonttitle=\bfseries\sffamily\scshape\small,
    coltitle=white,
    colbacktitle=mycolor!70!black,
    enhanced,
    attach boxed title to top left={xshift=2mm, yshifttext=-1mm, yshift=-2mm},
    boxed title style={
        outer arc=2mm,
        arc=1.5mm, 
        colframe=mycolor!65!black,
        colback=mycolor!70!black,
    },
    width=\textwidth,
    arc=3mm,
    boxrule=0.8pt,
  }
}

\title{Bottom-up Policy Optimization: Your Language Model Policy Secretly Contains Internal Policies}

\author{
\textbf{Yuqiao Tan}\textsuperscript{1,2}\thanks{Equal contribution.}
\hspace{0.6em}
\textbf{Minzheng Wang}\textsuperscript{1,2}\footnotemark[1]
\hspace{0.6em}
\textbf{Shizhu He}\textsuperscript{1,2}\thanks{Corresponding author.}
\hspace{0.6em}
\textbf{Huanxuan Liao}\textsuperscript{1,2}
\\
\textbf{Chengfeng Zhao}\textsuperscript{1,2}
\hspace{0.6em}
\textbf{Qiunan Lu}
\hspace{0.6em}
\textbf{Tian Liang}\textsuperscript{3}
\hspace{0.6em}
\textbf{Jun Zhao}\textsuperscript{1,2}
\hspace{0.6em}
\textbf{Kang Liu}\textsuperscript{1,2}
\\
\textsuperscript{1}Institute of Automation, Chinese Academy of Sciences
\\
\textsuperscript{2}University of Chinese Academy of Sciences
\\
\textsuperscript{3}Tencent AI Lab
\\
\texttt{tanyuqiao2025@ia.ac.cn}
\quad
\texttt{wangminzheng2023@ia.ac.cn}
\quad
\texttt{shizhu.he@nlpr.ia.ac.cn}
}

\begin{document}
\maketitle
\begin{abstract}
 Existing reinforcement learning (RL) approaches treat large language models (LLMs) as a unified policy, overlooking their internal mechanisms. 
In this paper, we decompose the LLM-based policy into \textbf{Internal Layer Policies} and \textbf{Internal Modular Policies} via the Transformer's residual stream. 
Our entropy analysis of internal policy reveals distinct patterns: (1) universally, internal policies evolve from high-entropy exploration in early layers to deterministic refinement in the top layers; and (2) Qwen exhibits an explicit progressive reasoning structure, contrasting with the abrupt convergence in Llama.
Furthermore, we discover that optimizing internal layers induces feature refinement, forcing lower layers to capture high-level reasoning representations early.
Motivated by these findings, we propose \textit{\textbf{Bottom-up Policy Optimization (BuPO)}}, a novel RL paradigm that reconstructs the LLM's reasoning foundation from the bottom up by optimizing internal layers in early stages.
Extensive experiments on complex reasoning benchmarks 
demonstrate the effectiveness of BuPO. Our code is available \href{https://github.com/Trae1ounG/BuPO}{here}.
\end{abstract}

\section{Introduction}
% Reinforcement learning (RL) has emerged as a key driver in advancing the complex reasoning capabilities of large language models (LLMs)~\citep{ouyang2022training, jaech2024openai}. Notably, the success of DeepSeek-R1~\cite{guo2025deepseek} has solidified reinforcement learning with verifiable rewards (RLVR) as a potent post-training paradigm for enhancing language model policy across diverse domains~\cite{yang2025qwen3, team2025kimi, yu2025dapo}. However, existing RL approaches typically treat the LLM as a unified policy, optimizing it solely on the final output distribution while neglecting the intricate information flow evolving through its internal residual streams.
% Mechanistic interpretability tools mitigate the opacity of black-box LLMs by unveiling their internal logic~\citep{belrose2023eliciting, tan-etal-2025-neural, gupta2025llms}.
% Such insights are viewed as a crucial bridge between deeper understanding of internal behavior and principled algorithmic refinement.
Reinforcement learning (RL) has become central to improving the complex
reasoning abilities of large language models (LLMs)~\citep{ouyang2022training,
jaech2024openai}. The success of DeepSeek-R1~\citep{guo2025deepseek} further
highlights reinforcement learning with verifiable rewards (RLVR) as an
effective post-training paradigm for improving LLM reasoning across
domains~\citep{yang2025qwen3, team2025kimi, yu2025dapo}. However, existing RL
methods typically treat an LLM as a unified policy, optimizing only the final
output distribution while overlooking the evolving information flow in its
internal residual streams. Mechanistic interpretability provides tools for
exposing such internal behavior~\citep{belrose2023eliciting, gupta2025llms}, creating an opportunity to connect
internal dynamics with more principled policy optimization.
% To date, most existing RLVR research has predominantly focused on surface algorithmic design, such as reward construction~\cite{shao2025spurious, chen2025pass, chen2025seed, liu2025understanding} and entropy regularization~\cite{cui2025entropy, yu2025dapo, yang2025dcpo}.

\begin{figure}[!t]
\centerline{\includegraphics[width=\columnwidth]{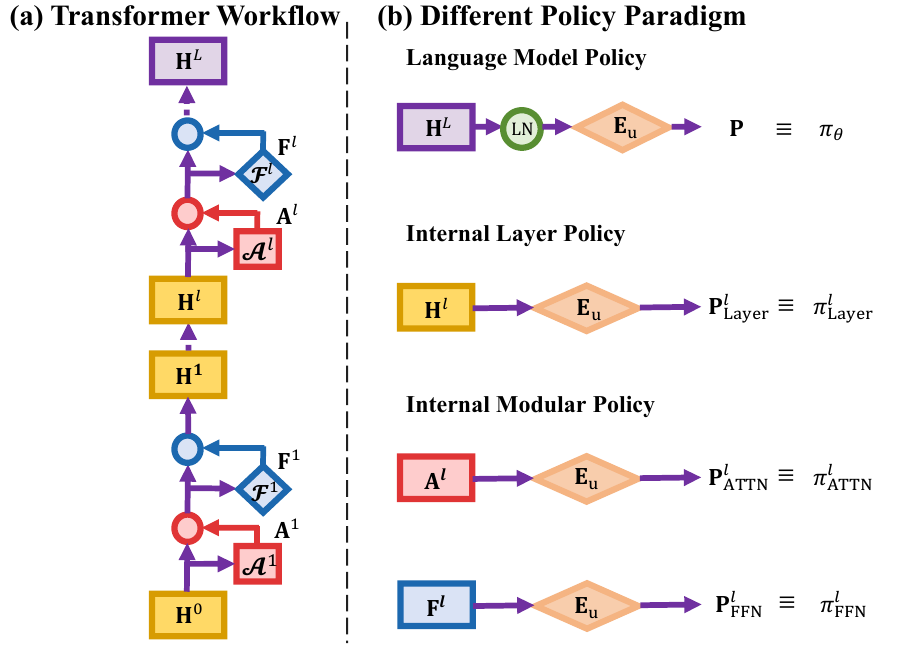}}
\caption{
(a) The residual stream within a Transformer flows from previous layer into self-attention  and feed-forward network (FFN) sequentially. (b) Any hidden states combined with the unembedding matrix $\mathbf{E}_\text{u}$ can be transformed into probability distribution $\mathbf{P}$ over the vocabulary space, which can be considered a policy. 
% \textbf{(c):} We surprisingly find that Qwen-series contains a progressive reasoning pattern in FFN, where it starts from exploration expansion to integrate middle layer knowledge into final prediction convergence.
}
\label{fig:intro}
\end{figure}
While recent works leverage attention mechanisms to improve RL
algorithms~\citep{li2025attention, liu2025attention}, they largely overlook the information latent in residual streams.
Logit-lens~\citep{nostalgebraist2020logitlens} offers initial insights by using
the unembedding matrix $\mathbf{E}_\text{u}$ to decode intermediate layer
representations into token space, revealing rich information that evolves across
layers and modules~\citep{gupta2025llms,
lindsey2025biology}. Moreover, prior studies have further elucidated the roles
of self-attention and feed-forward networks (FFNs) in shaping hidden
states~\citep{dai2022knowledge, yu2023neuron, jin2025massive}. Collectively,
these internal mechanisms offer a new perspective for algorithmic optimization.

In this paper, we investigate the evolution of language model policies across layers and modules to facilitate optimization and unravel complex internal reasoning mechanisms. 
Our formulation is grounded in two key insights. First, the residual stream naturally supports decomposition~\citep{zhang2025reinforcement, lindsey2025biology}, allowing us to isolate the individual roles of each layer and module (Figure~\ref{fig:intro}(a)).
Second, we conclude that the policy is intrinsically equivalent to the token distribution derived from the combination of hidden states $\mathbf{H}$ with the unembedding matrix $\mathbf{E}_{\text{u}}$.
 Based on these, we construct the \textbf{Internal Layer Policy}  $\pi_{\text{Layer}}^l$, which captures cumulative reasoning up to layer $l$, and the \textbf{Internal Modular Policy} $\pi_{\text{ATTN}}^l$ and $\pi_{\text{FFN}}^l$, which isolates the specific contributions of attention and FFN modules (Figure~\ref{fig:intro}(b)). This decomposition allows us to ask: \textit{How does internal reasoning evolve through the model?}

Through systematic analysis of commonly used Qwen and Llama series~\citep{meta2024llama32, yang2024qwen2,yang2025qwen3} based on \textbf{Internal Policy Entropy} in a policy-centric view, we uncover both universal and critical architectural differences:
 \textbf{(1) Consistent internal reasoning structure.} All models exhibit a universal reasoning structure: early layers maintain high entropy for exploring the solution space, while top layers converge to near-zero for final prediction~\citep{lindsey2025biology}. 
 \textbf{(2) Distinct internal reasoning pattern.} 
 Despite the shared trend, the pace of convergence differs significantly. Llama exhibits a sudden convergence only within the last three layers. In contrast, Qwen models demonstrate a progressive contraction, gradually reducing uncertainty throughout layers.
 To quantify these dynamics, we introduce \textbf{Entropy Change}. This metric shows that Llama exhibits increased internal entropy and tends to converge only in the final layer, whereas Qwen progressively leverages FFNs to expand exploration in lower layers, integrate parametric knowledge in intermediate layers, and consolidate predictions in upper layers.

These findings have profound implications for RL optimization: \textit{Since internal reasoning emerges from lower to higher, we can consider optimization from a bottom-up perspective.} 
We first validate this hypothesis with Internal Policy Alignment (IPA), revealing distinct training dynamics and a remarkable phenomenon of internal reasoning feature refinement. Specifically, the optimized lower layers capture high-level reasoning capabilities by early alignment, providing a robust foundation for subsequent internal reasoning.
Motivated by these insights, we propose \textit{\textbf{Bottom-up Policy Optimization (BuPO)}}, a novel RL paradigm that optimizes fine-grained internal layer policies during the early stages of training to effectively guide the language model policy. By doing so, BuPO reconstructs foundational reasoning abilities and achieves superior performance. Extensive experiments on complex reasoning benchmarks demonstrate the effectiveness of our approach and the unique training dynamics compared to conventional RL methods that optimize the policy as a whole.
% We hope our detailed analysis of language model policies provides valuable insights to improve RL algorithms.

In summary, this paper makes the following contributions: 
(1) We are the first to decompose an LLM policy into internal layer and modular policies, revealing their distinct roles in reasoning. (2) We reveal a universal exploration-to-convergence shift, distinguishing Qwen’s progressive reasoning from Llama’s abrupt convergence. Moreover, we uncover a feature refinement phenomenon where internal optimization drives lower layers to preemptively capture high-level features. (3) We propose BuPO to align internal policies from the bottom up, which reconstructs the foundational reasoning capabilities at lower layers, achieving superior performance on complex reasoning benchmarks.

% \begin{itemize}
%     \item \textbf{Internal Policy Decomposition:} First, we formally define and decompose language model policies into internal layer and modular policies, revealing their distinct roles in reasoning.
%     \item  \textbf{Internal Policy Entropy-based Analysis:} Building on this foundation, we uncover consistent yet distinct reasoning patterns across different model series, highlighting progressive reasoning structures in \texttt{Qwen}.
%     \item \textbf{Bottom-up Policy Optimization:} Inspired by these findings, we propose BuPO, a novel RL paradigm that optimizes internal policies early in training, achieving superior performance and unique training behaviors.
% \end{itemize}

\begin{figure*}[t!]
    \centering
    \subfigure{%
        \includegraphics[width=0.333\textwidth]{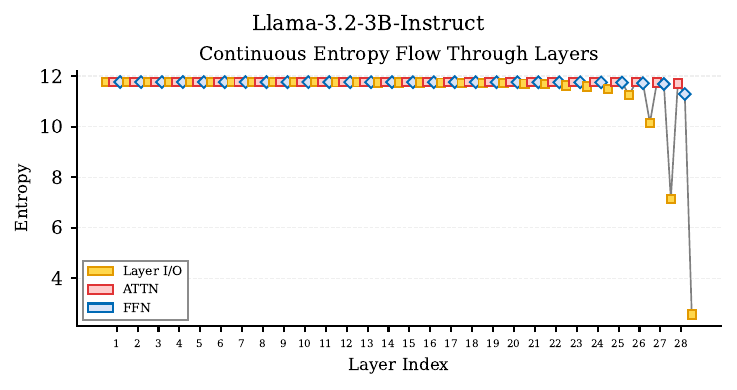}
    }%
    \subfigure{%
        \includegraphics[width=0.333\textwidth]{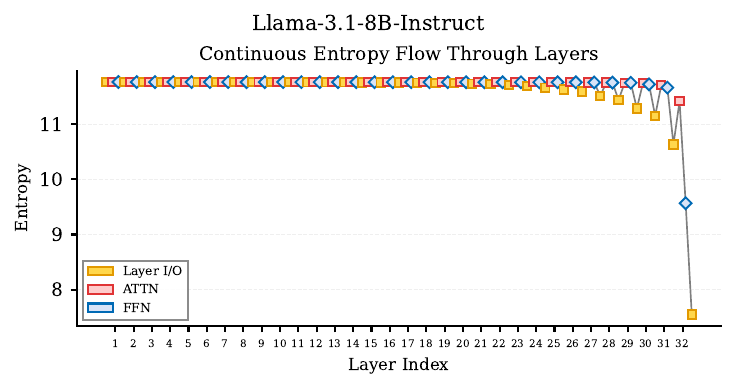}
    }%
    \subfigure{%
        \includegraphics[width=0.333\textwidth]{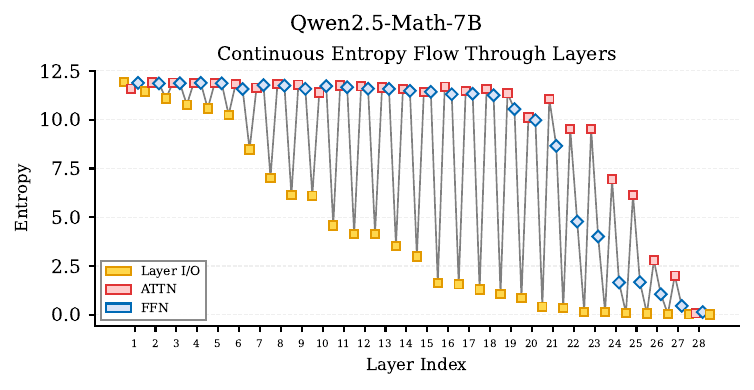}
    }%
    
    \subfigure{%
        \includegraphics[width=0.333\textwidth]{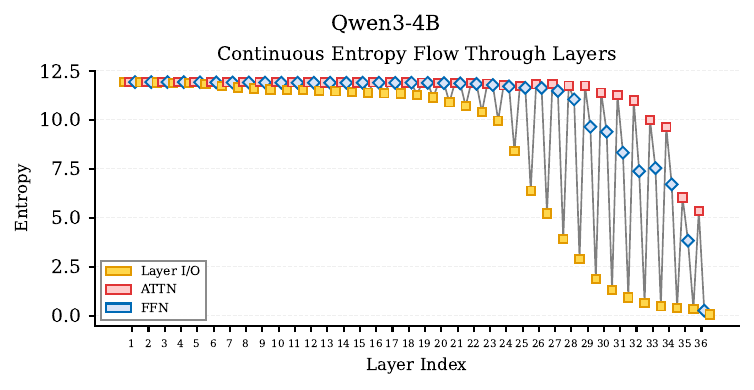}
    }%
    \subfigure{%
        \includegraphics[width=0.333\textwidth]{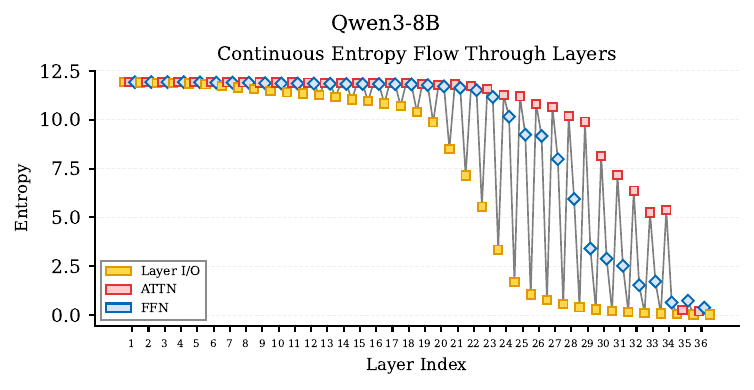}
    }%
     \subfigure{%
        \includegraphics[width=0.333\textwidth]{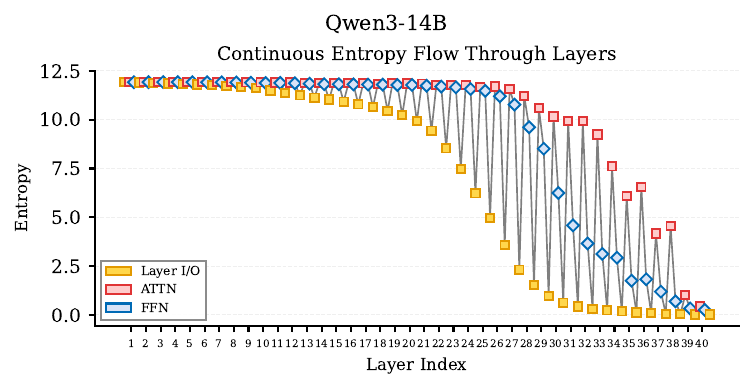}
    }%
    \caption{Continuous entropy dynamics of internal policy.
    % for different models. 
    The residual stream flows from $\mathbf{H}^{l-1}$ into $\mathbf{A}^l$, $\mathbf{F}^l$, then to the next layer $\mathbf{H}^l$.}
    \label{fig:interal_entropy_flow}
\end{figure*}

\section{Preliminary}
\label{sec:pre}
In this section, we aim to introduce the basic definitions that help us to understand the decomposition of the  language model policy. 
\subsection{The Residual Stream in Transformer}
\label{sec:transformer}

Transformer-based language models~\citep{vaswani2017attention} form the foundation of modern LLMs~\citep{brown2020language}. A decoder-only Transformer consists of $L$ stacked layers, each containing a multi-head self-attention (MHSA) module and a feed-forward network (FFN) module.

Following~\citet{zhang2025reinforcement}, we formalize the forward process from input to output. Given an input sequence $\mathbf{x} = [x_1, x_2, \dots, x_T]$, the model produces a probability distribution $\mathbf{P}$ over the vocabulary $V$ with $N$ tokens. Let $\mathbf{H}^{(2l-2)} \in \mathbb{R}^{T\times d_\text{model}}$ denote the hidden state input to the $l$-th layer, where $T$ is the sequence length and $d_\text{model}$ is the hidden dimension. The initial state is $\mathbf{H}^{(0)}$ projected by $\mathbf{E}$, where $\mathbf{E} \in \mathbb{R}^{N \times d_\text{model}}$ is the embedding matrix.

% Each layer forwards sequentially through attention and FFN:
% \begin{equation}
% \begin{aligned}
% \mathbf{A}^{l} &= \text{MHSA}(\texttt{LN}(\mathbf{H}^{(2l-2)})), \mathbf{H}^{(2l-1)}= \mathbf{H}^{(2l-2)} + \mathbf{A}^{l}, \\
% \mathbf{F}^{l} &= \text{FFN}(\texttt{LN}(\mathbf{H}^{(2l-1)})), \mathbf{H}^{(2l)} = \mathbf{H}^{(2l-1)} + \mathbf{F}^{l},
% \end{aligned}
% \end{equation}
% where $\texttt{LN}(\cdot)$ denotes layer normalization, and $\mathbf{A}^{l}, \mathbf{F}^{l}$ represent the attention and FFN outputs, respectively.

% Each layer forwards sequentially through attention and FFN: the attention output $\mathbf{A}^{l} = \mathrm{MHSA}(\mathrm{LN}(\mathbf{H}^{(2l-2)}))$ is added back via residual connection to give $\mathbf{H}^{(2l-1)} = \mathbf{H}^{(2l-2)} + \mathbf{A}^{l}$, followed by the FFN output $\mathbf{F}^{l} = \mathrm{FFN}(\mathrm{LN}(\mathbf{H}^{(2l-1)}))$ with another residual to yield $\mathbf{H}^{(2l)} = \mathbf{H}^{(2l-1)} + \mathbf{F}^{l}$, where $\mathrm{LN}(\cdot)$ denotes layer normalization.

Each layer forwards sequentially through attention and FFN with residual connections: $\mathbf{H}^{(2l-1)} = \mathbf{H}^{(2l-2)} + \mathbf{A}^{l}$ and $\mathbf{H}^{(2l)} = \mathbf{H}^{(2l-1)} + \mathbf{F}^{l}$, where $\mathbf{A}^{l} = \mathrm{MHSA}(\mathrm{LN}(\mathbf{H}^{(2l-2)}))$ and $\mathbf{F}^{l} = \mathrm{FFN}(\mathrm{LN}(\mathbf{H}^{(2l-1)}))$ are the attention and FFN outputs, and $\mathrm{LN}(\cdot)$ denotes layer normalization.

After $L$ layers, the final hidden states are projected to  vocabulary logits:
\begin{equation}
\mathbf{P} = \text{softmax}(\texttt{LN}(\mathbf{H}^{(2L)})\mathbf{E}_{\text{u}}^\text{T}),
\end{equation}
where $\mathbf{E}_{\text{u}} \in \mathbb{R}^{N \times d_{\text{model}}}$ is the unembedding matrix, and $\mathbf{P} \in \mathbb{R}^{T \times N}$ denotes the output distribution.

\subsection{Reinforcement Learning for Language Model Policy}
\label{sec:rl}

Language model generation can be formulated as a token-level Markov Decision Process (MDP). At each step $t$, the state $s_t = [\mathbf{q};\mathbf{o}_{<t}]$ consists of the input and generated tokens so far. The language model policy $\pi_\theta(\cdot | s_t)$ samples the next token $o_t$ from vocabulary $V$, transitioning to $s_{t+1}=[s_t;o_t]$. 
% Generation terminates upon producing \texttt{[eos]} or reaching the budget.
To optimize the policy, we maximize:
\begin{equation}
\begin{aligned}
    \label{eq:rl_objective}
    \mathcal{J}(\pi_\theta&) =
    {\mathbb{E}}_{{\mathbf{q} \sim \mathcal{Q},\mathbf{o} \sim \pi_\theta(\cdot|\mathbf{q}) }}\\
    &[R(\mathbf{q}, \mathbf{o})  - \beta \mathbb{D}_{KL}[\pi_\theta(\cdot|\mathbf{q})) || \pi_{\text{ref}}(\cdot|\mathbf{q})],
\end{aligned}
\end{equation}
where $R(\mathbf{q}, \mathbf{o})=\sum_{t=1}^{|\mathbf{o}|}r(s_t, o_t)$ is the return~\citep{sutton1998reinforcement} and $\pi_{\text{ref}}$ is a reference policy. We adopt sparse rewards where $r_t = 0$ for $t < n$ and $r_n \in [0, 1]$ indicates task success.
Following~\citet{hu2025openreasonerzeroopensourceapproach}, we assume $\beta=0$.
% throughout this paper.

\paragraph{\textbf{Policy Optimization.}} We adopt GRPO~\citep{shao2024deepseekmath}, which samples a group of responses $\{\mathbf{o}_1, \dots, \mathbf{o}_G\}$ per question and estimates advantages as $\hat{A}_{i,t}=\frac{R_i - \operatorname{mean}(\mathbf{R})}{\operatorname{std}(\mathbf{R})}$:
\begin{equation}
\label{eq:grpo}
    \begin{aligned}   
    &\mathcal{J}_{\text{GRPO}}(\pi_\theta) =   
    \mathbb{E}_{\mathbf{q}\sim \mathcal{Q},\{\mathbf{o_i}\}^G_{i=1}\sim\pi_{\theta_\text{old}}(\cdot|\mathbf{q})}
    \frac{1}{G}\sum^G_{i=1}\frac{1}{|\mathbf{o}_i|}\\
    &\sum_{t=1}^{|\mathbf{o_i}|}\left\{\min \left[r_{i,t}\hat{A}_{i,t}, \text{clip}(r_{i,t}, 1-\epsilon, 1+\epsilon)\hat{A}_{i,t}\right]\right\},
\end{aligned}
\end{equation}
where $r_{i,t}=\frac{\pi_\theta(o_{i,t}|s_{i,t})}{\pi_{\theta_\text{old}}(o_{i,t}|s_{i,t})}$ is the importance ratio.

\begin{figure*}[t!]
    \centering
    \includegraphics[width=\textwidth]{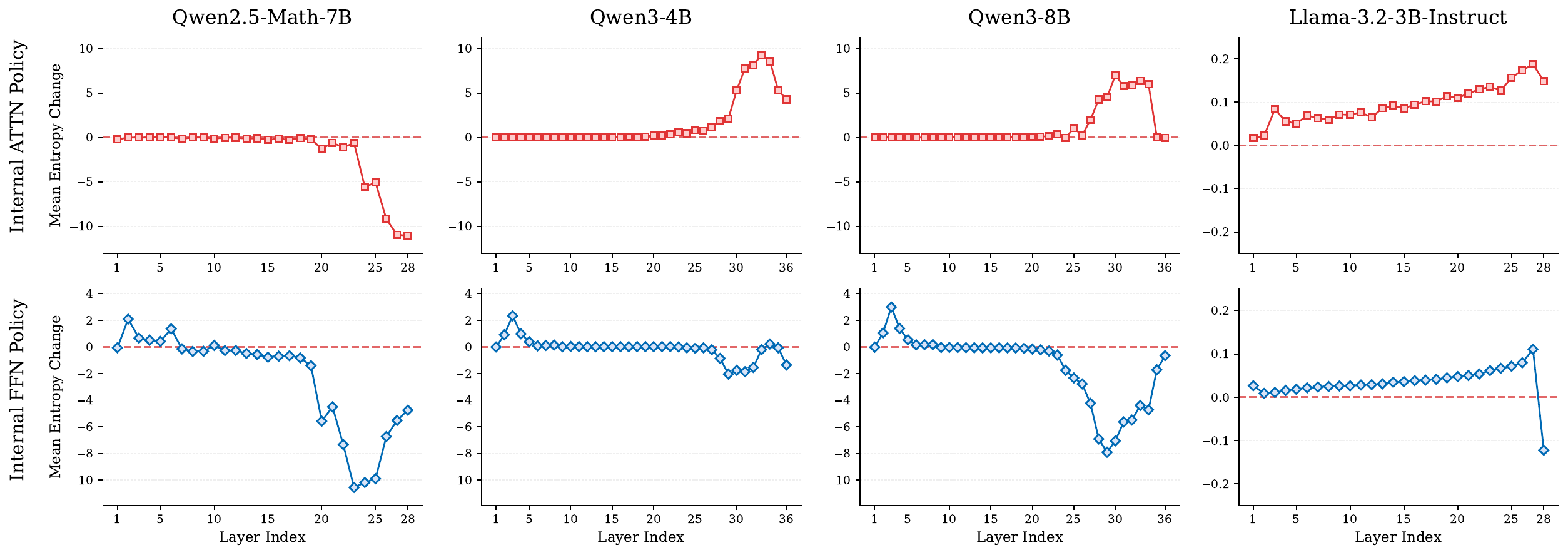}
    \caption{Entropy change dynamics of internal policy. The entropy change $\Delta H^l$ across layers represents the uncertainty of current policy's hidden exploration space. $\Delta H^l > 0$ indicates increased exploration, $\Delta H^l \approx 0$ signifies exploitation of existing knowledge, and $\Delta H^l < 0$ suggests convergence within the reasoning process.}
    \label{fig:modular_entropy}
\end{figure*}

\section{Language Model Policy Secretly Contains Internal Policies}
\label{sec:internal_in_language}
In this section, we introduce our key insight: \textit{the language model policy secretly contains internal policies}. We provide implementation details of this section in Appendix~\ref{app:sec3_imple}.
\subsection{Definition of Internal Policy}
\paragraph{\textbf{Residual Stream.}}
In the residual stream of Transformer, the input to any layer equals the sum of all preceding outputs plus the original embedding.
% Let $\mathbf{A}^l = \mathcal{A}^l(\mathbf{X}^{l}_\text{attn})$ and  $\mathbf{F}^l = \mathcal{F}^l(\mathbf{X}^l_\text{ffn})$ denote attention and FFN transformations (excluding normalization). 
The hidden states satisfy:
\begin{equation}
\mathbf{H}^{l}=\mathbf{H}^{(2l)} = \mathbf{H}^{(0)} + \sum_{i=1}^{l} \mathbf{A}^i + \sum_{j=1}^{l} \mathbf{F}^j, 
\end{equation}
where we denote $\mathbf{H}^l$ for simplicity as the output hidden states of layer $l$, $\mathbf{H}^{(2l)}$. According to this, the output of final layer can be regarded as the combination of previous hidden states by $\mathbf{H}^{L}=\mathbf{H}^{0}+\sum^{L}_{i=1} \mathbf{A}^i+\sum^j_{j=1}\mathbf{F}^j$.

\paragraph{\textbf{Internal Policy.}} During RL, we sample the next token $o_t$ from the final layer's probability distribution, i.e., $\pi_\theta \equiv \mathbf{P}=\text{softmax}(\texttt{LN}(\mathbf{H}^L)\mathbf{E}_\text{u}^\text{T})$. We propose that each internal hidden states can be combined with  $\mathbf{E}_\text{u}$ to produce a policy. Specifically, we focus on two granularities. \textbf{Internal Layer Policy} refers to using hidden states from each layer $\mathbf{H}^l$ to combine with $\mathbf{E}_\text{u}$, and \textbf{Internal Modular Policy} integrates $\mathbf{E}_u$ with states from specific module:
\begin{equation}
    \pi^l_\text{Layer}\equiv \mathbf{P}^l_{\text{Layer}}=\text{softmax}(\mathbf{H}^l\mathbf{E}_\text{u}^{\text{T}}),
\end{equation}
\begin{equation}
    \pi^l_{\text{Module}} \equiv 
    \begin{cases}
        \mathbf{P}^l_{\text{ATTN}}=\text{softmax}(\mathbf{A}^l\mathbf{E}_\text{u}^{\text{T}}),  \hfill \text{for ATTN}& \\
        \mathbf{P}^l_{\text{FFN}} =\text{softmax}(\mathbf{F}^l\mathbf{E}_\text{u}^{\text{T}}).  \hfill \text{for FFN} &
    \end{cases}
\end{equation}

Each component contributes to the final policy through the residual stream. For instance, $\mathbf{H}^L = \mathbf{H}^l + \mathbf{S}^{l+1}$, where $\mathbf{S}^{l+1} = \sum^L_{i=l+1}\mathbf{A}^i + \sum^L_{j=l+1}\mathbf{F}^j$ represents contributions from subsequent layers. Hence, understanding these internal components is essential for unraveling \textit{how internal reasoning emerges and evolves through the model}.

\subsection{Internal Policy Entropy Dynamics}
\label{sec:inter_policy_entropy_dy}
In contrast to prior logit-lens approaches~\citep{nostalgebraist2020logitlens} that decode internal states into discrete tokens, we adopt a policy-centric perspective where internal probability distributions are treated as policies. We employ \textbf{Entropy} as our primary probing metric, motivated by its strong correlation with policy behavior~\citep{cui2025entropy,cheng2025reasoning}. We define \textbf{Internal Policy Entropy} as:
\begin{equation}
    H_{\text{Layer}}^l=-\sum^{|V|}_{j=1}\mathbf{P}^l_{\text{Layer},j}\cdot \text{log}(\mathbf{P}^l_{\text{Layer},j}),
\end{equation}
where $|V|$ denotes the vocabulary size and we can obtain $H_{\text{FFN}}^l$ and $H_{\text{ATTN}}^l$ in the same way.

\paragraph{\textbf{Continuous Entropy Dynamics.}}
Figure~\ref{fig:interal_entropy_flow} shows that internal policy entropy dynamics exhibit consistent patterns across models: early layers maintain high entropy for exploration of the search space, while top layers converge to near-zero entropy. This aligns with findings that lower layers capture semantic information while higher layers aggregate and refine these representations to drive final decision-making~\citep{lindsey2025biology}.

While the overall entropy pattern is consistent across models, the fine-grained transition dynamics vary. To isolate intrinsic patterns from normalization and residual effects~\citep{he2016deep,zhang2019root},  we introduce \textbf{Entropy Change}, which measures the incremental information gain within a single internal policy and is defined as:
\begin{equation}
    \Delta H^l=H^l_{\text{Output}}-H^l_{\text{Input}},
\end{equation}
where the entropy change is defined as the difference between the internal policy entropy at a module's input and output. This metric reveals how the exploration space evolves as information propagates through the specific module. 
% Specifically, $\Delta H^l_{\text{FFN}} > 0$ indicates an expansion of exploration, $\Delta H^l_{\text{FFN}} \approx0$ suggests internal knowledge integration, and $\Delta H^l_{\text{FFN}} < 0$ reflects prediction convergence during the reasoning process.

\begin{figure*}[t!]
    \centering
    \includegraphics[width=1\textwidth]{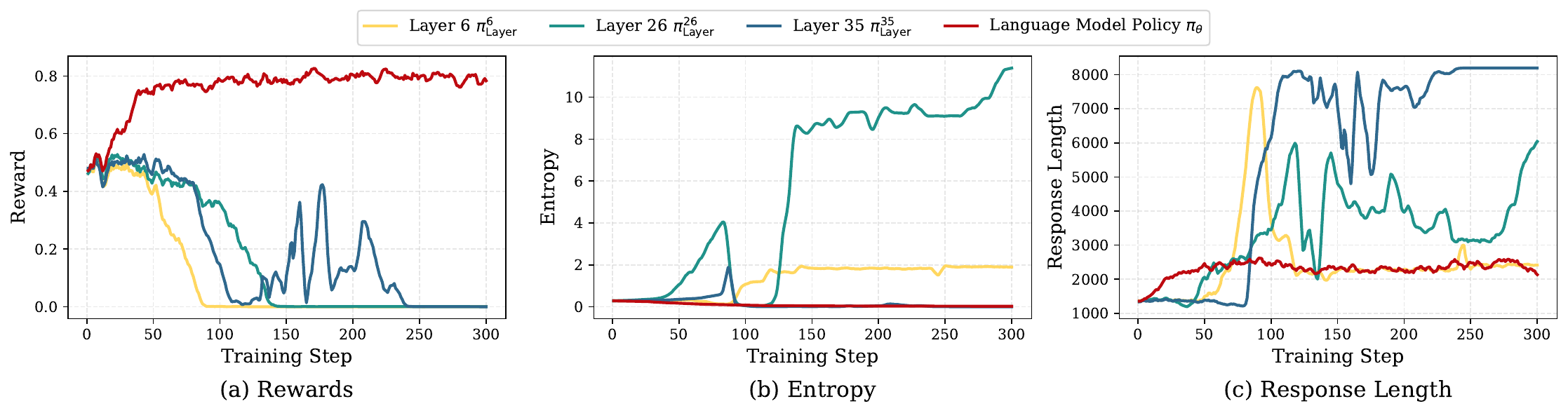}
    \caption{Training dynamics of internal policy. Effects of varying the optimized policy on (a) reward, (b) entropy of language model policy $\pi_\theta$, (c) response length. The backbone model is \texttt{Qwen3-4B}.}
    \label{fig:internal_train_dy}
\end{figure*}

\paragraph{\textbf{Entropy Change Dynamics of Attention vs. FFN.}}  \textit{Self-attention} modules~\citep{vaswani2017attention} are widely regarded as central to model reasoning, particularly for integrating task-relevant contextual information~\citep{jin2025massive, liu2025attention}. The upper panel of Figure~\ref{fig:modular_entropy} reveals a clear and model-dependent pattern in the entropy change of self-attention. Specifically, Qwen3 models exhibit consistently positive entropy change across layers ($\Delta H^l_{\text{ATTN}} > 0$), indicating sustained expansion of the exploration space during reasoning. In contrast, \texttt{Qwen2.5-Math-7B} shows uniformly negative entropy change, suggesting contraction and earlier convergence driven by attention. Llama models display a weaker but still positive trend, reflecting more conservative exploration dynamics. 

\begin{figure*}[t!]
    \centering
    \includegraphics[width=1\textwidth]{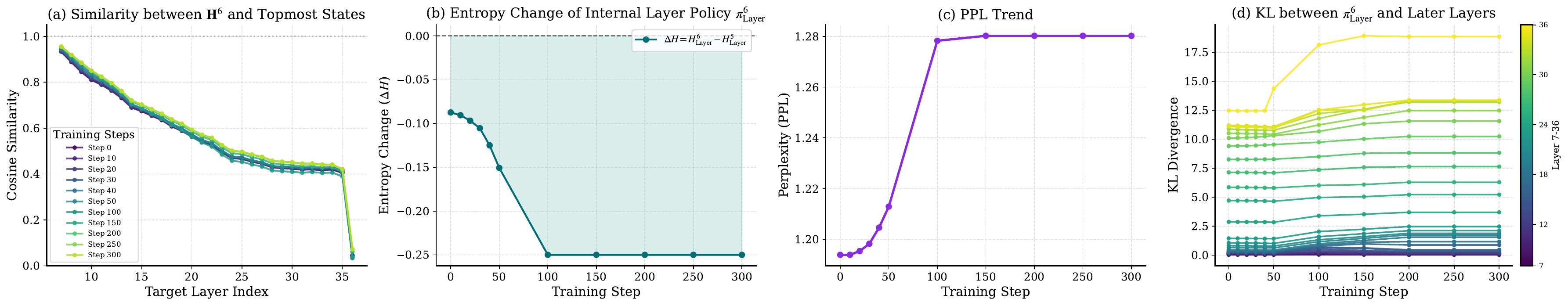}
    % \caption{Analysis of internal policy optimization. (a) Similarity between the hidden states of optimized layer 6 and the higher layers. (b) Entropy Change $\Delta H^6_\text{Layer}$ of the optimized $\pi_\text{Layer}^{6}$. (c) The PPL trend of the language model policy $\pi_\theta$. The backbone model is \texttt{Qwen3-4B}.}
    \caption{Analysis of internal policy optimization. (a) Similarity between the hidden states of optimized layer 6 and the higher layers. (b) Entropy change $\Delta H^6_\text{Layer}$ of the optimized internal layer policy $\pi_\text{Layer}^{6}$. (c) The PPL trend of $\pi_\theta$. (d) KL divergence trends between the optimized $\pi_\text{Layer}^{6}$ and later layer policies. The backbone model is \texttt{Qwen3-4B}.}
    \label{fig:analysis_combined}
\end{figure*}

\label{sec:ffn_entropy}
Moreover, the \textit{FFN} modules are widely regarded as the key-value memories of parametric knowledge~\citep{geva2021transformer, dai2022knowledge}. The lower panel of Figure~\ref{fig:modular_entropy} reveals clear and systematic differences in FFN entropy dynamics across model families. For the Llama models~\citep{meta2024llama32}, FFN entropy remains consistently positive but weak across almost all layers, with convergence occurring only at the final layer.
% This pattern indicates sustained but weak exploration throughout the FFN in Llama.  
% which aligns with prior observations that Llama models benefit less from post-training and require additional mid-training interventions to improve reasoning behavior~\citep{wang2025octothinker}. 

By contrast, the Qwen models~\citep{yang2024qwen2, yang2025qwen3} exhibit a pronounced hierarchical entropy structure in the FFN, following a clear three-stage progression. Taking \texttt{Qwen3-4B} as an example, the lower FFN layers (layers 1–6) show increased entropy ($\Delta H^l_{\text{FFN}} > 0$), corresponding to expanded exploration at the onset of reasoning. This is followed by a broad middle region (layers 7–26) where $\Delta H^l_\text{FFN}\approx0$, indicating stable information integration through retrieval and reuse of parametric knowledge
% encoded in knowledge neurons
~\citep{dai2022knowledge}. In the upper layers (layers 27–36) where $\Delta H^l_{\text{FFN}} < 0$, reflects gradual convergence toward the final prediction. 
\section{Internal Policy Alignment}
\label{sec:internal_policy_optim}
Building on the preceding analysis, we observe a gradual emergence of
internal reasoning: layer-induced policy distributions exhibit distinct roles
and entropy dynamics for different models. This motivates a bottom-up question: \textit{can an
internal policy distribution be refined before optimizing the full
policy?} We study this through \textit{Internal Policy Alignment} (IPA), an
advantage-weighted clipped surrogate for a selected internal layer policy:
\begin{equation}
\label{eq:ipa}
\begin{aligned}
&\mathcal{J}_{\text{IPA}}(\pi_\theta, \pi_{\text{Layer}}^l) =
\mathbb{E}_{\mathbf{q}\sim \mathcal{Q},
\{\mathbf{o}_i\}_{i=1}^{G}\sim
\pi_{\theta_{\text{old}}}(\cdot|\mathbf{q})} 
\frac{1}{G}\sum_{i=1}^{G}
\\ &
\frac{1}{|\mathbf{o}_i|}\sum_{t=1}^{|\mathbf{o}_i|}
\min\left[
\rho^l_{i,t}\hat{A}_{i,t},
\operatorname{clip}(\rho^l_{i,t},1-\epsilon,1+\epsilon)\hat{A}_{i,t}
\right],
\end{aligned}
\end{equation}
where
$\rho^l_{i,t} =
\frac{
\pi_{\text{Layer}}^l(o_{i,t}\mid \mathbf{q},\mathbf{o}_{i,<t})
}{
\pi_{\text{Layer,old}}^l(o_{i,t}\mid \mathbf{q},\mathbf{o}_{i,<t})
}$.
The sample process and $\hat{A}_{i,t}$ are the same as in Eq.~\ref{eq:grpo}. Thus, $\rho^l_{i,t}$ is not an importance correction for
the rollout distribution, but an internal policy update ratio that limits the
change of the selected layer-induced distribution on reward-driven tokens.
IPA transfers reward feedback to intermediate representations
while retaining the stability of clipped policy updates. Implementation details
are in Appendix~\ref{app:imple_internal_optimization}.

\paragraph{\textbf{Different Training Dynamics of Internal Policy.}} As shown in Figure~\ref{fig:internal_train_dy}, distinct patterns emerge across optimization of different internal layer policies.
For the penultimate layer policy $\pi_\text{Layer}^{35}$, entropy shows minor fluctuations before aligning with $\pi_\theta$. However, it suffers from repetition causing excessively long responses.
% , likely because final decision-making is confined to the last layer~\citep{gupta2025llms}.
In contrast, the last integration region policy $\pi_\text{Layer}^{26}$ exhibits unstable and increased entropy. And the last exploration region policy $\pi_\text{Layer}^6$ maintains stable entropy growth, with response lengths converging closer to $\pi_\theta$. 

\paragraph{\textbf{Analysis of Internal Policy Optimization.}}
We further investigate the mechanism behind internal policy alignment. We find
that it induces clear feature refinement in internal states. Taking
$\pi_\text{Layer}^6$ as an example, Figure~\ref{fig:analysis_combined}(a)
shows that the optimized $\mathbf{H}^6$ becomes increasingly similar to final
layer representations, suggesting that early layers acquire higher-level
reasoning features before full policy optimization. Meanwhile, the entropy
change in Figure~\ref{fig:analysis_combined}(b) shows progressive convergence
of the optimized internal policy. 
Figure~\ref{fig:analysis_combined}(c,d) reveals the trade-off of
internal policy alignment: while the KL trends show that $\pi_\text{Layer}^6$
remains well aligned with later-layer policies during the early stage, this
alignment deteriorates with prolonged optimization and coincides with PPL
degradation, suggesting that limited internal alignment steps are preferable.
% Figure~\ref{fig:analysis_combined}(d) further
% shows that the KL divergence between $\pi_\text{Layer}^6$ and later layer
% policies follows a coherent layer-wise trend, indicating that IPA reshapes the internal policy toward the downstream layer trajectory rather than only changing an isolated layer. Finally, the PPL trend
% in Figure~\ref{fig:analysis_combined}(c) reveals a trade-off: excessive
% alignment causes performance degradation, making limited internal alignment
% steps preferable.
\begin{table*}[t!]
    \centering
    \small
    
    \resizebox{1\textwidth}{!}{
    \begin{tabular}{lccccc}
        \toprule
        \textbf{Methods}  &  \textbf{AMC (Avg@16)} & \textbf{MATH500 (Avg@16)} & \textbf{AIME24 (Avg@32)}  & \textbf{AIME25 (Avg@32)}  & \textcolor{white}{11}\textbf{Average}\textcolor{white}{11}\\
        \midrule
        \texttt{Qwen3-4B} \\
        \quad Vanilla & 67.66 & 80.29 & 23.20 & 18.60 & 47.44  \\
        \quad PPO & 77.03 & \underline{83.64} & \underline{32.60} & 27.60 & \underline{55.22}  \\
        \quad Reinforce++ & 63.44 & 80.63 & 17.40 & 18.65 & 45.03  \\
        \quad RLOO & \underline{77.66} & 82.73 & 30.83 & 24.79 & 54.00 \\
        \quad GRPO & 76.88 & 82.41 & 32.19 & \underline{28.85} & 55.08  \\
        % \rowcolor{table-blue!66}\quad BuPO ($\pi_\text{Layer}^6 \rightarrow \pi_\theta, s_\text{inter}=30$) & 
        \rowcolor{table-blue!66}\quad BuPO &         \textbf{81.09}\rlap{\textbf{\scriptsize \textcolor{darkgreen}{+4.21}}} & \textbf{84.90}\rlap{\textbf{\scriptsize \textcolor{darkgreen}{+2.49}}} & \textbf{36.88}\rlap{\textbf{\scriptsize \textcolor{darkgreen}{+4.69}}} & \textbf{31.15}\rlap{\textbf{\scriptsize \textcolor{darkgreen}{+2.30}}} & \textbf{58.51}\rlap{\textbf{\scriptsize \textcolor{darkgreen}{+3.43}}} \\
        \midrule
        \texttt{Qwen3-8B} \\
        \quad Vanilla & 67.34 & 80.46 & 26.98 & 19.17 & 48.49  \\
        \quad PPO & \underline{87.03} & 86.20 & 37.81 & 22.60 & 58.41  \\
        \quad Reinforce++ & 82.66 & 86.05 & 41.77 & 31.15 & 60.41  \\
        \quad RLOO & 86.41 & 87.32 & 46.67 & 33.02 & 63.36  \\
        \quad GRPO & 85.94 & \textbf{88.05} & \underline{49.48} & \underline{33.54} & \underline{64.23}  \\
        % \rowcolor{table-blue!66}\quad BuPO ($\pi_\text{Layer}^6 \rightarrow \pi_\theta,s_\text{inter}=20$) &         
        \rowcolor{table-blue!66}\quad BuPO &\textbf{89.22}\rlap{\textbf{\scriptsize \textcolor{darkgreen}{+3.28}}} & \underline{87.76} & \textbf{54.06}\rlap{\textbf{\scriptsize \textcolor{darkgreen}{+4.58}}} & \textbf{34.38}\rlap{\textbf{\scriptsize \textcolor{darkgreen}{+0.76}}} & \textbf{66.36}\rlap{\textbf{\scriptsize \textcolor{darkgreen}{+2.13}}}  \\
        \midrule
        \texttt{Llama-OctoThinker-3B-Base} \\
        \quad Vanilla & 1.24 & 5.26 & 0.21 & 0.00 & 1.68  \\
        \quad PPO & 22.19 & 43.23 & \underline{1.04} & \underline{0.31} & 16.69 \\
        \quad Reinforce++ & 9.38 & 11.59 & 0.00 & 0.10 & 5.27  \\
        \quad RLOO & \underline{27.03} & 41.93 & \textbf{2.19} & 0.21 & 17.84  \\
        \quad GRPO & \textbf{27.50} & \underline{46.07} & 0.63 & 0.10 & \underline{18.58}  \\
        % \rowcolor{table-blue!66}\quad BuPO ($\pi_\text{Layer}^{27} \rightarrow \pi_\theta, s_\text{inter}=20$)&
        \rowcolor{table-blue!66}\quad BuPO & \textbf{27.50}\rlap{\textbf{\scriptsize \textcolor{darkgreen}{+0.00}}} & \textbf{49.79}\rlap{\textbf{\scriptsize \textcolor{darkgreen}{+3.72}}} & 0.63\rlap{\textbf{\scriptsize \textcolor{darkgreen}{+0.00}}} & \textbf{0.42}\rlap{\textbf{\scriptsize \textcolor{darkgreen}{+0.32}}} & \textbf{19.59}\rlap{\textbf{\scriptsize \textcolor{darkgreen}{+1.01}}}  \\
        \midrule
         \texttt{Llama-OctoThinker-8B-Base} \\
        \quad Vanilla & 4.53 & 9.84 & 0.52 & 0.10 & 3.75 \\
        \quad PPO & 31.72 & 56.97 & 1.56 & 1.04 & 22.82  \\
        \quad Reinforce++ & 34.69 & \underline{59.55} & \textbf{7.72} & \underline{3.75} & \underline{26.43}  \\
        \quad RLOO & 27.66 & 55.97 & 3.54 & 1.56 & 22.18  \\
        \quad GRPO & \underline{34.84} & 56.89 & 2.50 & 2.19 & 24.11  \\
        % \rowcolor{table-blue!66}\quad BuPO ($\pi_\text{Layer}^{31} \rightarrow \pi_\theta, s_\text{inter}=20$)& 
        \rowcolor{table-blue!66}\quad BuPO & 
        \textbf{37.66}\rlap{\textbf{\scriptsize \textcolor{darkgreen}{+2.82}}} & \textbf{62.05}\rlap{\textbf{\scriptsize \textcolor{darkgreen}{+5.16}}} & \underline{4.69}\rlap{\textbf{\scriptsize \textcolor{darkgreen}{+2.19}}} & \textbf{6.77}\rlap{\textbf{\scriptsize \textcolor{darkgreen}{+4.58}}} & \textbf{27.79}\rlap{\textbf{\scriptsize \textcolor{darkgreen}{+3.68}}}  \\
        \bottomrule
    \end{tabular}
    }
    \caption{Avg@$K$ results on MATH500, AMC23, AIME24 and AIME25. \textbf{Bold} and \underline{underlined} denote the best and second best.}
    \label{tab:main_exp}
\end{table*} 

\section{Bottom-up Policy Optimization}
\label{sec:bupo}
All prior RL methods for LLMs optimize $\pi_\theta$ in a holistic manner~\citep{ouyang2022training,rafailov2023direct, guo2025deepseek}. 
In Section~\ref{sec:internal_in_language}, we decompose the final output $\mathbf{H}^L$ into an intermediate representation $\mathbf{H}^l$ and the subsequent residual contribution $\mathbf{S}^{l+1}$. 
This decomposition suggests that aligning the internal policy $\pi_{\text{Layer}}^{l}$ associated with $\mathbf{H}^l$ may facilitate alignment of the overall policy $\pi_\theta$. 
We empirically support this intuition in Section ~\ref{sec:internal_policy_optim}, where we optimize the internal policy alone and observe pronounced feature refinement in lower layers. 
These findings further motivate a bottom-up alignment strategy: by aligning internal policies first, we explore whether the overall policy can be guided to reason more effectively.

To this end, we propose \textbf{\textit{Bottom-up Policy Optimization (BuPO)}}, which sequentially optimizes the internal layer policies $\pi_\text{Layer}^l$ followed by the language model policy $\pi_\theta$:
% The overall training objective is:
% \begin{equation}
%     \mathcal{J}_{\text{BuPO}}(\pi_\theta,\pi_\text{Layer}^l)=
%     \begin{cases}
%         \mathcal{J}_{\text{InterGRPO}}(\pi_\theta,\pi_\text{Layer}^l), & s_{\text{cur}} \leq s_{\text{inter}} \\
%         \mathcal{J}_{\text{GRPO}}(\pi_\theta), &s_\text{cur} > s_\text{inter}
%     \end{cases}
% \end{equation}
\begin{equation}
\mathcal{J}_{\text{BuPO}}
=
\begin{cases}
\mathcal{J}_{\text{IPA}}, & s_{\text{cur}} \le s_{\text{inter}},\\
\mathcal{J}_{\text{GRPO}}, & s_{\text{cur}} > s_{\text{inter}},
\end{cases}
\end{equation}
where $s_\text{cur}$ denotes the current training step, $s_\text{inter}$ the training steps of the internal layer policy.

\begin{figure*}[h!]
    \centering
    \includegraphics[width=1\textwidth]{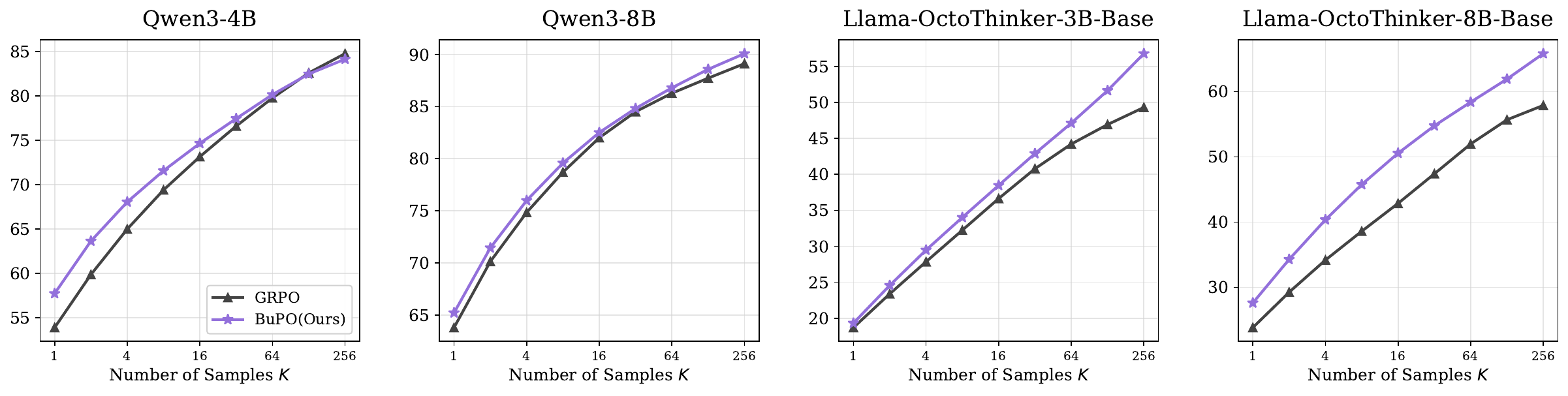}
    \caption{Average Pass@$K$ results on MATH500, AMC23, AIME24 and AIME25.}
    \label{fig:passk_main}
\end{figure*}
% \textbf{Training Setup.} Specifically, we focus on the Qwen3 series, which maintain a stable EIC pattern, using \texttt{Qwen3-4B} and \texttt{Qwen3-8B}~\citep{yang2025qwen3}. Meanwhile, we select \texttt{Llama-OctoThinker-3B-Base} and \texttt{Llama-OctoThinker-8B-Base} from the Llama series, as these models demonstrate improved training behavior after mid-training based on \texttt{Llama-3.2-Base}~\citep{wang2025octothinker}.
% We present the detailed training algorithm in Appendix~\ref{app:bupo_imple} and training setup in Appendix~\ref{app:imple_bupo}.

\textbf{Training Setup.} Using Qwen~\citep{yang2025qwen3} and Llama~\citep{grattafiori2024llama, wang2025octothinker} backbones, 
% we select the last FFN layer with a positive exploration signal ($\Delta H^l_\text{FFN} >0$) as $\pi_\text{Layer}^l$. 
we assign $\pi_\text{Layer}^6$ for both \texttt{Qwen3-4B} and \texttt{Qwen3-8B}, $\pi_\text{Layer}^{27}$ for \texttt{Llama-OctoThinker-3B-Base}, and $\pi_\text{Layer}^{31}$ for \texttt{Llama-OctoThinker-8B-Base}. For BuPO, we use the entropy-indicated target layer as the internal policy to optimize, and discuss this choice through the layer-policy ablation in Section~\ref{sec:ablation}.
 Refer to Appendix~\ref{app:imple_bupo} for detailed setup.

\textbf{Evaluation Setup.} We evaluate BuPO against several RL baselines, including GRPO, PPO~\citep{sutton1998reinforcement}, Reinforce++~\citep{hu2025reinforce++} and RLOO~\citep{ahmadian2024back}. The benchmarks cover: MATH~\citep{lightman2023let}, AMC23~\citep{AMC23}, AIME24, and AIME25~\citep{AIME24,AIME25}.  Due to high output variance in reasoning tasks, we report Avg@$K$ (Pass@1 averaged over $K$ outputs). For AIME24/25, we set $K=32$, and for others $K=16$. Additionally, we evaluate an unbiased Pass@$K$ metric for a comprehensive evaluation. Detailed evaluation setups are in Appendix~\ref{app:imple_bupo}.

\begin{figure}[!t]
    \centering
    \subfigure{
        \includegraphics[width=0.48\columnwidth]{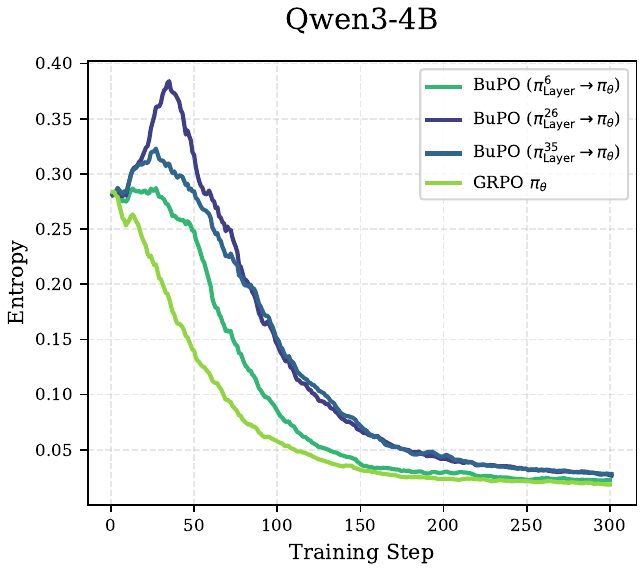}
    }%
    \subfigure{
        \includegraphics[width=0.48\columnwidth]{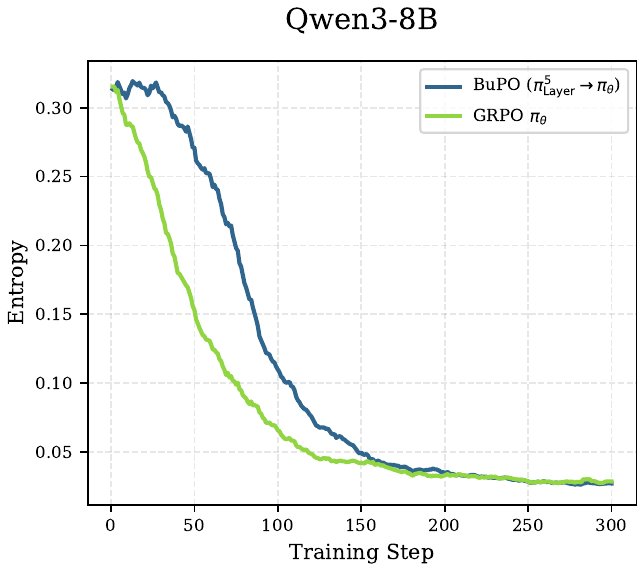}
    }\\[-2ex]
    \subfigure{
        \includegraphics[width=0.48\columnwidth]{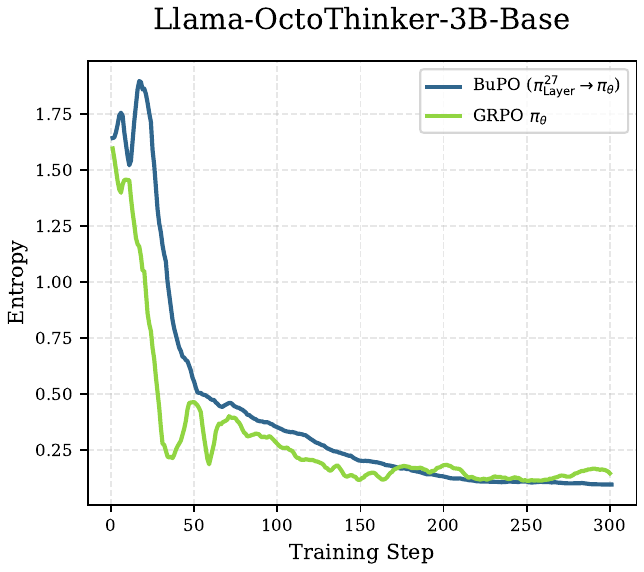}
    }%
    \subfigure{
        \includegraphics[width=0.48\columnwidth]{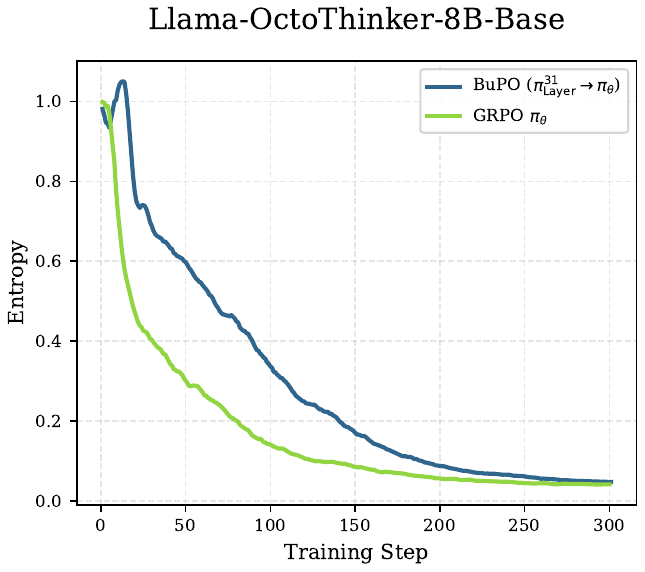}
    }
    \caption{Entropy dynamics during training with GRPO and BuPO with different internal policy.}
    \label{fig:train_entropy_dyn}
\end{figure}

\subsection{Main Results}
\label{sec:main_results}
% We report the Avg@k results in Table~\ref{tab:main_exp}. 
% Surprisingly,
As shown in Table~\ref{tab:main_exp}, BuPO consistently improves over baselines across models and benchmarks. On \texttt{Qwen3-4B}, BuPO achieves gains of 4.69 and 2.30 points on AIME24 and AIME25, respectively, with similar improvements on \texttt{Qwen3-8B} (+4.58 and +0.76). The Llama series further validates this trend, with average improvements of 1.01 points on \texttt{Llama-OctoThinker-3B-Base} and 3.68 points on \texttt{Llama-OctoThinker-8B-Base}.  We further evaluate BuPO's generalization on out-of-domain benchmarks spanning general reasoning and coding in Appendix~\ref{app:ood}.
These results demonstrate that aligning internal layer policies provides a consistent and effective training signal for improving reasoning.
% As shown in Table ~\ref{tab:main_exp},
% adopting the bottom-up perspective in BuPO leads to consistent Avg@$K$ improvements over RL baseline algorithms across benchmarks and models, achieving superior average performance. 
% On \texttt{Qwen3-4B}, BuPO yields gains of 4.69 points on AIME24 and 2.30 points on AIME25 compared to GRPO. 
% Similarly, the \texttt{Qwen-8B} model shows improvements of 4.58 points on AIME24 and 0.76 points on AIME25. 
% The Llama series models exhibit a similar optimization trend under BuPO, achieving an average improvement of 1.01 points on \texttt{Llama-OctoThinker-3B-Base} and 3.68 points on \texttt{Llama-OctoThinker-8B-Base}.
% Overall, these results suggest that aligning internal policies in the early stages of training can effectively guide the language model policy toward improved reasoning.

For a  comprehensive evaluation, we report the averaged Pass@$K$ results of BuPO and GRPO, with $K$ ranging from 1 to 256. 
In Figure~\ref{fig:passk_main}, BuPO consistently achieves a favorable trade-off across a wide range of $K$. 
On \texttt{Qwen3-8B}, BuPO attains the best performance for all $K$ values, while on \texttt{Qwen3-4B}, the only exception occurs at $K=256$. 
For all Llama models, BuPO achieves the best results across all $K$ values, yielding gains of 7.48 points on \texttt{Llama-OctoThinker-3B-Base} and 7.93 points on \texttt{Llama-OctoThinker-8B-Base} in Pass@256. 
These results further indicate the IPA effectively enhances the reasoning capacity of $\pi_\theta$. 
% We provide detailed Pass@$K$ performance in Figure~\ref{fig:passk_app}.

\subsection{Analysis}
\label{sec:analysis}
\paragraph{\textbf{Training Dynamics of BuPO.}} We further visualize the training dynamics of BuPO.
 % following the settings in Sec.~\ref{sec:main_results}. 
 As shown in Figure~\ref{fig:train_entropy_dyn}, by aligning the internal layer policy at an early stage, all models exhibit enhanced entropy exploration initially.
 For Qwen models, optimizing layer 6 maintains stable exploration, consistent with the later layer-policy ablation. 
 For Llama, we observe increased entropy during the bottom alignment stage, indicating that feature refinement in lower layers effectively provides a larger exploration space for $\pi_\theta$.

\subsection{Ablation Study}
\label{sec:ablation}
% \textit{\textbf{(1) Effect of Bottom Optimization Steps.}}
% \paragraph{Effect of Bottom Optimization Steps.}

% % \paragraph{\textbf{Ablation of Bottom Optimization Steps.}}  
% % Sec.~\ref{sec:internal_policy_optim} reveals that as the optimization steps of the internal layer policy progress, the language model policy eventually collapses.
% We further conduct an ablation study of how $s_\text{inter}$ affects the performance of BuPO. As shown in the upper panel of Table~\ref{tab:ablation}, as the bottom alignment steps increase, the performance of the language model policy drops dramatically. This aligns with the finding in Figure~\ref{fig:analysis_combined} and further supports our conclusion in Section~\ref{sec:analysis} that only moderate bottom optimization can boost overall policy learning effectively.
% % \paragraph{\textbf{Ablation of optimized internal policy.}}

\paragraph{Effect of Bottom Optimization Steps.}
We further study how the number of bottom optimization steps $s_\text{inter}$
affects BuPO. As shown in Table~\ref{tab:ablation_s_inter}, moderate internal
policy alignment consistently improves over GRPO on \texttt{Qwen3-4B}.
Specifically, setting $s_\text{inter}=30$ improves the average score from
55.08 to 58.51, while increasing it to 50 or 70 steps causes a drop.
These results are consistent with Figure~\ref{fig:analysis_combined}:
limited bottom optimization promotes internal policy alignment, whereas
excessive alignment destabilizes the language model policy and harms final
performance.
% We further study how the number of bottom optimization steps $s_\text{inter}$
% affects BuPO. As shown in Table~\ref{tab:ablation_s_inter}, moderate internal
% policy alignment consistently improves over GRPO on both \texttt{Qwen3-4B} and \texttt{Qwen3-8B}.
% For \texttt{Qwen3-4B}, setting $s_\text{inter}=30$ improves the average score from
% 55.08 to 58.51, while increasing it to 50 or 70 steps causes a sharp drop. A
% similar trend appears on \texttt{Qwen3-8B}, where $s_\text{inter}=20$ improves the
% average score from 64.23 to 66.36, but $s_\text{inter}=50$ degrades it to
% 57.14. These results are consistent with Figure~\ref{fig:analysis_combined}:
% limited bottom optimization promotes internal policy alignment, whereas
% excessive alignment destabilizes the language model policy and harms final
% performance.

% \paragraph{ Effect of Optimized Internal Policy.}
% % \textit{\textbf{(2) Effect of Optimized Internal Policy.}}
% We further investigated the impact of the optimized internal policy, focus on the region boundaries in Section~\ref{sec:inter_policy_entropy_dy}. As shown in the lower panel of Table~\ref{tab:ablation}, BuPO variants employing different $\pi_\text{Layer}^l$ exhibit superior performance over GRPO with a fixed $s_\text{inter}$. The entropy dynamics in Figure~\ref{fig:train_entropy_dyn} indicate that the alignment of $\pi_\text{Layer}^{26}$ induces an entropy spike, leading to significant exploration expansion, aligning with its strong performance. The selection and investigation of internal policies remain promising directions for future research.

\paragraph{Effect of Optimized Internal Policy.}
We further study the effect of choosing different optimized internal policies
$\pi_\text{Layer}^l$. As shown in Table~\ref{tab:ablation_layer_policy}, the
best-performing layer varies across backbones: $\pi_\text{Layer}^{6}$ achieves
the best performance on both \texttt{Qwen3-4B} and \texttt{Qwen3-8B}, while
$\pi_\text{Layer}^{26}$ and $\pi_\text{Layer}^{30}$ perform best on
\texttt{OctoThinker-3B} and \texttt{OctoThinker-8B}, respectively. Despite this
variation, these best-performing layers share a common pattern in
Figure~\ref{fig:modular_entropy}: each corresponds to the last layer with a
positive FFN exploration signal $\Delta H_\mathrm{FFN}^l > 0$, i.e., the final layer before the FFN entropy dynamics shift from exploration to integration or convergence. 
For clarity, we refer to this entropy-indicated transition point as the \textbf{Boundary Layer}. Across all four backbones, optimizing the Boundary Layer achieves the best average performance, suggesting that the entropy-change pattern provides a useful practical signal for choosing the target internal policy.
% This provides a practical guideline for selecting which internal layer to optimize.
% In addition, Figure~\ref{fig:train_entropy_dyn} shows that different optimized internal
% policies lead to distinct entropy dynamics during training, further explaining
% why the choice of $\pi_\text{Layer}^l$ can substantially affect final
% performance. 
We analyze the computational cost of entropy-based inspection in Appendix~\ref{app:preliminary_internal_analysis_cost}, showing that its overhead is small compared with downstream policy training.

\section{Related Work}
\paragraph{\textbf{Reinforcement Learning with Verifiable Rewards.}}
% Reinforcement learning (RL) has proven effective for enhancing LLMs
% , most notably through Reinforcement Learning from Human Feedback (RLHF), which aligns model outputs with human preferences
% ~\citep{sutton1998reinforcement,rafailov2023direct,jaech2024openai,wang2025adaptive}.
Recently, Reinforcement Learning with Verifiable Rewards (RLVR) has gained traction for its ability to foster LLM reasoning using rule-based rewards~\citep{guo2025deepseek,yang2025qwen3,wang2025adaptive,tan2025zero,wang2026mitigating,tan2026p,wang2026breaking,liao2026resadapt}. 
In this work, we shift the focus to internal policies and propose bottom-up policy optimization, which directly optimizes internal layer policies in early training stages. This targeted optimization refines internal reasoning representations and ultimately leads to improved performance.

\paragraph{\textbf{Interpretability of LLMs.}}
Interpretability studies have opened LLM black boxes by analyzing how models
reason, store knowledge, and route information internally~\citep{tan2025neural, gupta2025llms, hu2025affects},
especially through self-attention~\citep{zhou2024role, jin2025massive} and
FFN~\citep{dai2022knowledge, meng2022locating}.  In this work, we conduct a systematic analysis of hidden states from a policy-centric perspective,  revealing structured internal reasoning patterns that motivate internal policy alignment which refines intermediate features before full policy optimization.
% Interpretability tools mitigate the opacity of black-box LLMs by revealing their internal logic. A substantial body of prior work investigates how LLMs  reason and memory~\citep{yu2023neuron, tan-etal-2025-neural, lindsey2025biology,gupta2025llms,hu2025affects}, with particular emphasis on the self-attention~\citep{zhou2024role,jin2025massive} and the feed-forward network~\citep{dai2022knowledge, meng2022locating,geva2023dissecting}. Insights into these internal mechanisms provide a new perspective for algorithmic optimization~\citep{li2025attention, liu2025attention}. In this work, we conduct a systematic analysis of internal hidden states from a policy-centric perspective and identify various internal reasoning patterns across models. 
% Motivated by these observations, we propose bottom-up policy optimization, which refines internal features and leads to a effective RL algorithm.

\begin{table}[t!]
    \centering
    \resizebox{1\columnwidth}{!}{
    \begin{tabular}{lccccc}
        \toprule
        \textbf{Methods} & \textbf{AMC} & \textbf{MATH500} & \textbf{AIME24} & \textbf{AIME25} & \textbf{Average} \\
        \midrule
        \multicolumn{6}{l}{\texttt{Qwen3-4B}} \\
        GRPO & 76.88 & 82.41 & 32.19 & 28.85 & 55.08 \\
        BuPO ($s_\text{inter}=30$) & \textbf{81.09} & \textbf{84.90} & \textbf{36.88} & \textbf{31.15} & \textbf{58.51} \\
        BuPO ($s_\text{inter}=50$) & 62.66 & 79.00 & 14.17 & 12.60 & 42.11 \\
        BuPO ($s_\text{inter}=70$) & 14.14 & 25.20 & 0.21 & 0.00 & 9.89 \\
        % \midrule
        % \multicolumn{6}{l}{\texttt{Qwen3-8B}} \\
        % GRPO & 85.94 & 88.05 & 49.48 & 33.54 & 64.23 \\
        % BuPO ($s_\text{inter}=20$) & \textbf{89.22} & \textbf{87.76} & \textbf{54.06} & \textbf{34.38} & \textbf{66.36} \\
        % BuPO ($s_\text{inter}=50$) & 78.98 & 85.40 & 37.08 & 27.08 & 57.14 \\
        \bottomrule
    \end{tabular}
    }
    \caption{Ablation of $s_\text{inter}$ with fixed layer policy $\pi_\text{Layer}^6$.}
    \label{tab:ablation_s_inter}
\end{table}

\begin{table}[t!]
    \centering
    \resizebox{1\columnwidth}{!}{
    \begin{tabular}{lccccc}
        \toprule
        \textbf{Layer Policy} & \textbf{AMC} & \textbf{MATH500} & \textbf{AIME24} & \textbf{AIME25} & \textbf{Average} \\
        \midrule
        \multicolumn{6}{l}{\texttt{Qwen3-4B}} \\
        \textbf{Boundary Layer} ($\pi_\text{Layer}^{6}$) & 81.09 & \underline{84.90} & \textbf{36.88} & \textbf{31.15} & \textbf{58.51} \\
        $\pi_\text{Layer}^{26}$ & \underline{81.56} & 84.40 & \underline{35.42} & \underline{30.52} & 57.98 \\
        $\pi_\text{Layer}^{35}$ & \textbf{82.66} & \textbf{85.14} & \underline{35.42} & 30.12 & \underline{58.34} \\
        \midrule
        \multicolumn{6}{l}{\texttt{Qwen3-8B}} \\
        \textbf{Boundary Layer} ($\pi_\text{Layer}^{6}$) & \textbf{89.22} & \textbf{87.76} & \textbf{54.06} & \textbf{34.38} & \textbf{66.36} \\
        $\pi_\text{Layer}^{21}$ & 75.55 & 84.40 & 32.60 & 21.67 & 53.56 \\
        
        $\pi_\text{Layer}^{35}$ & \underline{83.75} & \underline{86.51} & \underline{42.81} & \underline{29.58} & \underline{60.66} \\
        \midrule
        \multicolumn{6}{l}{\texttt{OctoThinker-3B}} \\
        \textbf{Boundary Layer} ($\pi_\text{Layer}^{26}$) & \textbf{27.50} & \textbf{49.79} & \textbf{0.63} & \textbf{0.42} & \textbf{19.59} \\
        $\pi_\text{Layer}^{16}$ & \underline{16.17} & \underline{11.40} & 0.00 & 0.00 & \underline{6.89} \\
        $\pi_\text{Layer}^{6}$ & 7.80 & 9.60 & \underline{0.21} & \underline{0.10} & 4.43 \\
        \midrule
        \multicolumn{6}{l}{\texttt{OctoThinker-8B}} \\
        \textbf{Boundary Layer} ($\pi_\text{Layer}^{30}$) & \textbf{37.66} & \textbf{62.05} & \textbf{4.69} & \textbf{6.77} & \textbf{27.79} \\
        $\pi_\text{Layer}^{16}$ & 3.91 & \underline{12.20} & 0.00 & 0.00 & 4.03 \\
        $\pi_\text{Layer}^{6}$ & \underline{7.58} & 9.40 & 0.00 & 0.10 & \underline{4.27} \\
        \bottomrule
    \end{tabular}
    }
    
    \caption{Ablation of the optimized internal policy $\pi_\text{Layer}^l$. The \textbf{Boundary Layer} denotes the target layer selected for BuPO, corresponding to the last layer with positive FFN entropy change.}
    \label{tab:ablation_layer_policy}
\end{table}

\section{Conclusion}
In this paper, we decompose a language model policy into internal layer and
modular policies, revealing systematic reasoning patterns through entropy
analysis. We observe a broad transition from high-entropy exploration in early
layers to deterministic convergence in higher layers, with Qwen models showing a progressive reasoning structure contrasting with the abrupt convergence observed in
Llama. Motivated by these findings, we propose Bottom-up Policy Optimization
(BuPO), an RL paradigm that aligns internal layer policies during early
training. Extensive experiments on complex reasoning benchmarks demonstrate its
effectiveness. Further analysis shows that internal policy alignment refines
foundational reasoning features, thereby improving the reasoning capacity of the
overall policy.

\section*{Limitations}
While BuPO demonstrates consistent improvements by leveraging internal policy optimization, several practical limitations offer avenues for future research.
(1) Similar to existing RLVR methods, BuPO requires substantial computational resources for rollout generation, reward evaluation, and policy optimization. Our experiments were conducted on a single node with 8 NVIDIA A100 GPUs. Such training requirements may limit accessibility for researchers with constrained computational budgets, and improving the efficiency of RL training remains an important direction for the community.
(2) BuPO is studied in reasoning-oriented settings, where models often generate long reasoning trajectories before producing final answers. This increases both training and evaluation costs, especially when evaluating reasoning performance with large-sample metrics such as Pass@$K$ with 300 sampled responses.
Overall, we hope BuPO inspires the design of more effective RL algorithms that make use of internal model structures, and encourages closer connections between reinforcement learning and interpretability for reasoning language models.

\section*{Ethical considerations}
Our approach does not introduce ethical concerns. The datasets we used are public, and there are no privacy issues.

\bibliography{custom}

\appendix
\clearpage
\section{Detailed Experiment Settings}

\subsection{The Models for Experiments}
We summarize all models used in our analysis and experiments in Table~\ref{tab:model_detail}. These models are categorized into three types: Mix, Base, and Instruct.

\begin{table*}[!htb]
\centering

\resizebox{1\textwidth}{!}{
\begin{tabular}{@{}llll@{}}
\toprule
\textbf{Model} & \textbf{Huggingface} & \textbf{Type} & \textbf{Layers}\\
\midrule
Qwen3-4B & \url{https://huggingface.co/Qwen/Qwen3-4B} & Mix & 36\\
Qwen3-8B & \url{https://huggingface.co/Qwen/Qwen3-8B} & Mix & 36 \\
Qwen3-14B & \url{https://huggingface.co/Qwen/Qwen3-14B} & Mix & 40 \\
Qwen3-4B-Base & \url{https://huggingface.co/Qwen/Qwen3-4B-Base} & Base & 36\\
Qwen2.5-Math-7B & \url{https://huggingface.co/Qwen/Qwen2.5-Math-7B} & Base & 28\\
Qwen3-4B-Instruct-2507 & \url{https://huggingface.co/Qwen/Qwen3-4B-Instruct-2507} & Instruct & 36\\
Llama-3.2-3B-Instruct & \url{https://huggingface.co/meta-llama/Llama-3.2-3B-Instruct}& Instruct & 28 \\
Llama-3.1-8B-Instruct & \url{https://huggingface.co/meta-llama/Llama-3.1-8B-Instruct}& Instruct  & 32\\
Llama-OctoThinker-3B-Base & \url{https://huggingface.co/OctoThinker/OctoThinker-3B-Long-Base}& Base & 28\\
Llama-OctoThinker-8B-Base & \url{https://huggingface.co/OctoThinker/OctoThinker-8B-Long-Base}& Base & 32 \\
 DeepSeek-Math-7B-Base & \url{https://huggingface.co/deepseek-ai/deepseek-math-7b-base} & Base & 30\\
 DeepSeek-R1-Distill-Qwen-7B & \url{https://huggingface.co/deepseek-ai/DeepSeek-R1-Distill-Qwen-7B} & Instruct & 28\\
 
\bottomrule
\end{tabular}
}
\caption{Detailed information about the selected models is provided. "Mix" refers to models that support both thinking and non-thinking modes. "Base" denotes the pre-trained model only. "Instruct" indicates models that undergo further fine-tuning based on the Base model to enhance instruction-following capabilities.}
\label{tab:model_detail}
\end{table*}

\subsection{The Template for Experiments}
\label{sec:template}

We adopt the following template for all experiments involving Qwen models, building upon the Qwen-Math template used for Qwen2.5~\citep{yang2024qwen2} and the Qwen-Nothinking template for Qwen3.

\begin{figure}[h]
\begin{tcolorbox}[colframe=gray, colback=mycolor!5!white, coltitle=white, title=Qwen-Math Template]
\texttt{<|im\_start|>system} \\
Please reason step by step, and put your final answer within \texttt{\textbackslash boxed\{\}}. \texttt{<|im\_end|>} \\
\texttt{<|im\_start|>user} \\
\texttt{\{problem\}} \\
\texttt{<|im\_end|>} \\
\texttt{<|im\_start|>}assistant
\end{tcolorbox}
\caption{Prompt template for Qwen-math.}
\end{figure}

\begin{figure}[h]
\begin{tcolorbox}[colframe=gray, colback=mycolor!5!white, coltitle=white, title=Qwen3-NoThinking Template]
\texttt{<|im\_start|>system} \\
Please reason step by step, and put your final answer within \texttt{\textbackslash boxed\{\}}. \texttt{<|im\_end|>} \\
\texttt{<|im\_start|>user} \\
\texttt{\{problem\}} \\
\texttt{<|im\_end|>} \\
\texttt{<|im\_start|>}assistant \\
\texttt{<think>} \\
\\
\texttt{</think>}\\
\end{tcolorbox}
\caption{Prompt template for Qwen3 NoThinking mode.}
\end{figure}

For training the \texttt{Llama-OctoThinker} models, we adopt the original prompt in~\citet{wang2025octothinker} to ensure performance.

\begin{figure}
\begin{tcolorbox}[colframe=gray, colback=mycolor!5!white, coltitle=white, title=OctoThinker Template]
A conversation between User and Assistant. The user asks a question, and the Assistant solves it. The assistant first thinks about the reasoning process in the mind and then provides the user with the answer. User: You must put your answer inside \texttt{\textbackslash boxed\{\}} and Your final answer will be extracted automatically by the \texttt{\textbackslash boxed\{\}} tag. \\
\texttt{\{problem\}} \\
Assistant:
\end{tcolorbox}
\caption{Prompt template for OctoThinker.}

\end{figure}

\subsection{Implementation of Internal Policy Entropy Analysis}
\label{app:sec3_imple}
In this section, we detail the implementation of internal policy entropy analysis. Our primary objective is to extract internal hidden states during the forward pass. In the main experiments, we evaluate model-generated responses on the MATH test set \citep{hendrycks2021measuring}. Entropy is computed at the token level for each layer and module, and then averaged over all generated tokens. We find that the intrinsic reasoning patterns remain stable across different tasks, e.g., commonsense question answering~\citep{rein2024gpqa}.

The computation of internal policy entropy is illustrated in the pseudo-code below. We abstract the entropy computation as a function $\mathcal{H}(\cdot)$. Accordingly, the entropy change of the internal layer policy is defined as:
\begin{equation}
\Delta H^l_{\text{Layer}} = \mathcal{H}(\mathbf{H}^l) - \mathcal{H}(\mathbf{H}^{l-1}).
\end{equation}

For the two core Transformer submodules, the entropy changes are computed separately. Specifically, for the self-attention module, we define:
\begin{equation}
\Delta H^l_{\text{ATTN}} = \mathcal{H}(\mathbf{A}^l) - \mathcal{H}(\texttt{LN}(\mathbf{H}^{(2l-2)})),
\end{equation}

and for the feed-forward network (FFN), we define:
\begin{equation}
\Delta H^l_{\text{FFN}} = \mathcal{H}(\mathbf{F}^l) - \mathcal{H}(\texttt{LN}(\mathbf{H}^{(2l-1)})).
\end{equation}

These definitions allow us to quantify how each layer and submodule contributes to the evolution of the internal policy entropy.

\begin{figure*}[t]
\begin{tcolorbox}[
    title=Calculation of Internal Policy Entropy (PyTorch Implementation),
    colback=white,
    colframe=darkspringgreen!40,
    coltitle=gray!40!black,
    fonttitle=\bfseries,
    arc=1mm,
    boxrule=0.6mm,
    left=1mm,   
    right=1mm,   
    top=1mm,     
    bottom=1mm,  
   width=\textwidth
]
\ttfamily
\# Get layer hidden states by register hook \\
hidden\_state = get\_from\_hook()\\[6pt]
\# Compute logits in the same way as in the original forward pass \\
logits = self.model.lm\_head(hidden\_state) \\[6pt]
\# Apply softmax for normalization \\
probs = torch.softmax(logits, dim=-1) \\[6pt]
\# Apply log\_softmax for speedy computation \\
log\_probs = torch.log\_softmax(logits, dim=-1) \\[6pt]
\# Calculate internal layer policy \\
\textcolor{myentropy}{entropies} = -(probs * log\_probs).sum(dim=-1)
\end{tcolorbox}
\caption{The pseudocode of internal policy entropy calculation.}
\label{tab:pseudocode}
\end{figure*}

\paragraph{\textbf{Discussion: Comparison to the Logit Lens.}}
Notably, our definition of internal policy differs from the logit-lens~\citep{nostalgebraist2020logitlens}, particularly in how layer normalization (\texttt{LN}) is handled. To clarify this distinction, we provide a systematic comparison between the two formulations in Table~\ref{tab:definition_comparison}. 

Our definition adopts a policy-centric perspective, treating the internal hidden states as explicitly policy distribution. In contrast, the logit-lens is primarily designed to project hidden states into the discrete vocabulary space in order to inspect the most likely output tokens at intermediate layers. We intentionally omit \texttt{LN} based on empirical considerations: in our analysis experiments, incorporating \texttt{LN} leads to less stable entropy dynamics and weaker interpretability than those shown in Figures~\ref{fig:interal_entropy_flow} and~\ref{fig:modular_entropy}.

% Extensive analyses and experiments with BuPO further demonstrate that our formulation of internal policy serves as a robust interpretability tool for uncovering internal reasoning mechanisms in LLMs. 

Extensive analyses and experiments with BuPO further demonstrate that our formulation of internal policy serves as a robust interpretability tool for uncovering internal reasoning mechanisms in LLMs. As shown in Table~\ref{tab:ablation_ln}, we also provide an ablation on layer normalization, finding that removing LN yields consistently better performance, which validates our design choice of omitting it in BuPO.

\begin{table}[!h]
    \centering
    
    \resizebox{1\columnwidth}{!}{
    \begin{tabular}{lcccc}
    \toprule
    \textbf{Methods} & \textbf{AMC} & \textbf{MATH} & \textbf{AIME24} & \textbf{AIME25} \\
    \midrule
    \multicolumn{5}{l}{\texttt{Qwen3-4B}} \\
    BuPO-wo-LN (Ours) & \textbf{81.09} & \textbf{84.90} & \textbf{36.88} & \textbf{31.15} \\
    BuPO-w-LN & 80.08 & 84.80 & 32.29 & 30.00 \\
    \midrule
    \multicolumn{5}{l}{\texttt{Qwen3-8B}} \\
    BuPO-wo-LN (Ours) & \textbf{89.22} & \textbf{87.76} & \textbf{54.06} & \textbf{34.38} \\
    BuPO-w-LN & 86.41 & 87.40 & 47.81 & 34.21 \\
    \bottomrule
    \end{tabular}
    }
    \caption{Ablation on layer normalization in BuPO.}
    \label{tab:ablation_ln}
\end{table}

% \begin{table}[!h]
%     \centering
%     \caption{Ablation on layer normalization in BuPO.}
%     \label{tab:ablation_ln}
%     \resizebox{1\columnwidth}{!}{
%     \begin{tabular}{llcccc}
%     \toprule
%     \textbf{Model} & \textbf{Method} & \textbf{AMC} & \textbf{MATH} & \textbf{AIME24} & \textbf{AIME25} \\
%     \midrule
%     \multirow{2}{*}{\texttt{Qwen3-4B}}
%         & \textbf{BuPO-wo-LN (Ours)} & \textbf{81.09} & \textbf{84.90} & \textbf{36.88} & \textbf{31.15} \\
%         & BuPO-w-LN                                           & 80.08          & 84.80          & 32.29          & 30.00 \\
%     \midrule
%     \multirow{2}{*}{\texttt{Qwen3-8B}}
%         &\textbf{BuPO-wo-LN (Ours)} & \textbf{89.22} & \textbf{87.76} & \textbf{54.06} & \textbf{34.38} \\
%         & BuPO-w-LN                                           & 86.41          & 87.40          & 47.81          & 34.21 \\
%     \bottomrule
%     \end{tabular}
%     }
% \end{table}

\begin{table}[ht]
\centering
\resizebox{\columnwidth}{!}{\begin{tabular}{lcc}
\toprule
 & \textbf{Logit Lens} & \textbf{Ours} \\
\midrule
Perspective & Discrete Token & Policy Distribution\\[6pt]
Definition & 
$\texttt{LN}(\mathbf{H}^l)\mathbf{E}_\text{u}^\text{T}$ & 
$\text{softmax}(\mathbf{H}^l\mathbf{E}_\text{u}^\text{T})$ \\[6pt]
Trainable & \xmark &  \cmark \\[6pt]
\bottomrule
\end{tabular}}
\caption{Comparison between the logit lens and our definition}
\label{tab:definition_comparison}
\end{table}

\subsection{Implementation of Internal Policy Alignment}
\label{app:imple_internal_optimization}

In internal policy alignment, namely IPA, at each optimization step, we select a specific internal layer $l$ and optimize its internal policy $\pi_{\text{Layer}}^{l}$, defined as $\pi_{\text{Layer}}^{l} = \text{softmax}(\mathbf{H}^{l} \mathbf{E}_\text{u}^\text{T})$, where $\mathbf{H}^{l}$ is the hidden states at layer $l$. The gradient flow for internal policy optimization is determined by the residual structure of the Transformer, where the hidden state $\mathbf{H}^{l}$ is a function of all parameters from layer $1$ to $l$, but is independent of parameters in higher layers.

Formally, for any parameter $\theta_k$ in layer $k$, the gradient of the IPA loss with respect to $\theta_k$ can be expressed using the chain rule:
\begin{equation}
\frac{\partial J_{\text{IPA}}(\pi_{\text{Layer}}^{l})}{\partial \theta_k}
= \frac{\partial J_{\text{IPA}}}{\partial \pi_{\text{Layer}}^{l}}
\cdot \frac{\partial \pi_{\text{Layer}}^{l}}{\partial \mathbf{H}^{l}}
\cdot \frac{\partial \mathbf{H}^{l}}{\partial \theta_k}
\end{equation}

Due to the residual connections, $\frac{\partial \mathbf{H}^{l}}{\partial \theta_k} \neq 0$ only when $k \leq l$, and is zero otherwise. Thus, the gradients for different layers can be summarized as:
\begin{equation}
\frac{\partial J_{\text{IPA}}(\pi_{\text{Layer}}^{l})}{\partial \theta_k} =
\begin{cases}
\frac{\partial J_{\text{IPA}}}{\partial \pi_{\text{Layer}}^{l}}
\cdot \frac{\partial \pi_{\text{Layer}}^{l}}{\partial \mathbf{H}^{l}}
\cdot \frac{\partial \mathbf{H}^{l}}{\partial \theta_k}, & \text{if } k \leq l \\
0, & \text{if } k > l
\end{cases}
\end{equation}

This means that, during IPA optimization for layer $l$, only the parameters of layers $0$ through $l$ and unembedding matrix $\mathbf{E}_\text{u}$ are updated, while all higher layers ($k > l$) remain unaffected. This targeted gradient flow ensures that internal policy optimization provides direct supervision to the selected layer and all lower layers, strengthening foundational reasoning capabilities without interfering with higher-level representations.

\paragraph{\textbf{Training Setup.}} Based on the findings in Section~\ref{sec:internal_in_language},  we select \texttt{Qwen3-4B}~\citep{yang2025qwen3} in non-thinking mode for investigation. 
For the training set, we randomly sample 5k entries from \textit{DeepMath-103k}~\citep{he2025deepmath}. We train the models using the verl framework~\citep{sheng2025hybridflow}. The prompt batch size is 128, with 8 rollouts generated per prompt. The sampling temperature during training is set to 1.0, and the maximum context length is set to  9,216 tokens. We update the model with a mini-batch size of 32 for 300 steps and a learning rate of 1e-6. 
These stage-wise FFN entropy patterns suggest that transition points between exploration and later integration/convergence may provide informative candidates for internal policy optimization.
For \texttt{Qwen3-4B}, the region boundaries identified in Section~\ref{sec:ffn_entropy} lie at layers 6 and 26. Additionally, we focus on aligning the internal policy of the penultimate layer, which is critical for the final prediction. Accordingly, we compare the internal policies $\pi_\text{Layer}^6$, $\pi_\text{Layer}^{26}$, and $\pi_\text{Layer}^{35}$ optimized via IPA, against the overall policy $\pi_\theta$ optimized with GRPO~\citep{he2025deepmath}.

\subsection{Implementation of Main Experiments}
\label{app:imple_bupo}

% \textbf{Training setup.}  Specifically, we focus on the \texttt{Qwen3} series, which maintain a stable Exploration-Integration-Convergence pattern, using \texttt{Qwen3-4B} and \texttt{Qwen3-8B}. For comparison, we select \texttt{Llama-OctoThinker-3B-Base} and \texttt{Llama-OctoThinker-8B-Base} from the \texttt{Llama} series, as these models demonstrate improved RL training performance after additional mid-training based on \texttt{Llama-3.2-Base}~\citep{wang2025octothinker}. 

% \rowcolor{table-blue!66}\quad BuPO ($\pi_\text{Layer}^6 \rightarrow \pi_\theta, s_\text{inter}=30$) &

% \rowcolor{table-blue!66}\quad BuPO ($\pi_\text{Layer}^6 \rightarrow \pi_\theta,s_\text{inter}=20$) &

% \rowcolor{table-blue!66}\quad BuPO ($\pi_\text{Layer}^{27} \rightarrow \pi_\theta, s_\text{inter}=20$)&

% \rowcolor{table-blue!66}\quad BuPO ($\pi_\text{Layer}^{31} \rightarrow \pi_\theta, s_\text{inter}=20$)&

\textbf{Detailed Training Setup.} 
% Regarding the selection of $\pi_\text{Layer}$, we identify the last FFN layer exhibiting a positive exploration signal (i.e., $\Delta H^l_\text{FFN} >0$) to serve as $\pi_\text{Layer}^l$ in our main experiments. Consequently, we assign $\pi_\text{Layer}^6$ for both \texttt{Qwen3-4B} and \texttt{Qwen3-8B}, $\pi_\text{Layer}^{27}$ for \texttt{Llama-OctoThinker-3B-Base}, and $\pi_\text{Layer}^{31}$ for \texttt{Llama-OctoThinker-8B-Base}. The training steps for the internal layer policy are set to $s_\text{inter}=30$ for \texttt{Qwen3-4B} and $s_\text{inter}=20$ for the other LLMs. 
Specifically, we focus on the Qwen3 series, using \texttt{Qwen3-4B} and \texttt{Qwen3-8B}~\citep{yang2025qwen3}. Meanwhile, we select \texttt{Llama-OctoThinker-3B-Base} and \texttt{Llama-OctoThinker-8B-Base} from the Llama series, as these models demonstrate improved training behavior after mid-training based on \texttt{Llama-3.2-Base}~\citep{wang2025octothinker}. 
We set $s_\text{inter}=30$ for \texttt{Qwen3-4B} and $s_\text{inter}=20$ for the other LLMs.
We implement GRPO and other baseline algorithms using the veRL framework~\citep{sheng2025hybridflow}. Across all algorithms and model variants, we adopt a unified set of hyperparameters, as reported in Table~\ref{tab:rl_parameters}, and do not employ entropy regularization or KL-based losses. For PPO, the critic network is trained separately with a learning rate of $1\times10^{-5}$. All experimental results are averaged over three random seeds. 

\textbf{Evaluation Setup.} We use vLLM~\citep{kwon2023efficient} with temperature 1.0 and top\_p 1.0. 
This metric is defined as Pass@$K:=\mathbb{E}_{x\sim\mathcal{D}}\left[1-\binom{n-c}{K}/\binom{n}{K}\right]$,  where $c$ denotes the number of correct completions out of $n$ generated responses. To reduce evaluation variance on those datasets, we set $n=300$.

\subsection{The Cost of Internal Policy Analysis}
\label{app:preliminary_internal_analysis_cost}

The internal policy analysis is a lightweight one-time diagnostic used to identify the layer for bottom-up policy optimization. All experiments are conducted on a single node with 8 NVIDIA A100 GPUs. As shown in Figure~\ref{tab:pseudocode}, for each sample, the analysis only requires a standard forward pass: we register hooks to cache intermediate hidden states, project them with the shared unembedding matrix, and compute token-level entropy. Therefore, it introduces no backward pass, policy update or reference-model pass.

\paragraph{Time Cost.}
Taking \texttt{Qwen3-4B} as an example, the full RL training takes 13.22 hours, i.e., about 47,592 seconds. In contrast, the internal policy entropy analysis takes only 209.25 seconds. This corresponds to about $0.44\%$ of the RL training time, or an analysis-to-training ratio of approximately $1:227.4$. Such a small one-time cost is sufficient to recover stable internal reasoning patterns for selecting the optimized internal layer.

\paragraph{FLOPs Cost.}
Let $P$ denote the number of model parameters, $B$ the number of analyzed samples, and $S$ the average token count per sample. A typical RL step contains generation, policy forward, backward/update, and reference passes, costing roughly $10PS$ FLOPs per sample. For \texttt{Qwen3-4B}, with 300 steps and 1,024 samples per step, the total training cost is approximately $300 \times 1024 \times 10 \times 4\text{B} \times S
\approx 12288\,S \ \text{TFLOPs}$.
By comparison, the analysis requires only a single forward pass, about $2PBS$, plus entropy computation dominated by LM-head projections across internal layers. The total cost is approximately $17.6\,S$ TFLOPs, yielding an analysis-to-training FLOPs ratio of about $1:698.2$.

\begin{table}[t]
\centering
% \small
\resizebox{\columnwidth}{!}{
\begin{tabular}{lccc}
\toprule
\textbf{Metric} & \textbf{RL Training} & \textbf{Analysis} & \textbf{Ratio} \\
\midrule
Total Time & 13.22h ($\sim$47,592s) & 209.25s & $\sim$1:227.4 \\
Total FLOPs & $12288\,S$ TFLOPs & $17.6\,S$ TFLOPs & $\sim$1:698.2 \\
\bottomrule
\end{tabular}
}
\caption{Cost comparison between RL training and internal policy analysis on \texttt{Qwen3-4B}.}
\label{tab:cost_analysis}
\end{table}

Moreover, the analysis can be directly performed on the baseline model rollouts at step 0, requiring no additional rollout collection. These results show that internal policy analysis is substantially cheaper than downstream RL training while providing the necessary signal for practical layer selection.
% In Sec.~\ref{sec:internal_policy_optim} and Sec.~\ref{sec:bupo}, we conduct reinforcement learning (RL) experiments to evaluate the performance of the proposed algorithm. In this section, we describe the RL training setup in detail. We implement GRPO and other baseline algorithms using the veRL framework~\citep{sheng2025hybridflow}. Across all algorithms and model variants, we adopt a unified set of hyperparameters, as reported in Table~\ref{tab:rl_parameters}, and do not employ entropy regularization or KL-based losses. For PPO, the critic network is trained separately with a learning rate of $1\times10^{-5}$.

\begin{table}[!t]
\centering

\resizebox{\columnwidth}{!}{
\begin{tabular}{@{}ll@{}}
\toprule
\textbf{Hyperparameter} & \textbf{Value} \\
\midrule
Optimizer & AdamW \\
Policy learning rate & $1\text{e}^{-6}$ \\
Critic learning rate & $1\text{e}^{-5}$ (for PPO) \\
Training batch size & 128 prompts \\
Samples per prompt & 8 \\
Mini-batch size & 32 prompts \\
Policy updates per rollout & 16 \\
Max prompt length & 1024 tokens \\
Max response length & 7168 tokens (Qwen) / 3072 (Llama) \\
Rollout temperature & 1.0 \\
Clip range $\epsilon$ & 0.2 \\
\bottomrule
\end{tabular}
}
\caption{RL Hyperparameters}\label{train_hyperparameters}
\label{tab:rl_parameters}
% \vspace{-5pt}
\end{table}

\newpage
\begin{table*}[!t]
\centering

\setlength{\tabcolsep}{6pt}
\small
\begin{tabular}{lcccccccc}
\toprule
\textbf{Methods} & \textbf{Olympiad} & \textbf{Minerva} & \textbf{MMLU-Pro} & \textbf{GPQA-D} & \textbf{ARC-C} & \textbf{BBH} & \textbf{HumanEval} & \textbf{Avg} \\
\midrule
\multicolumn{9}{l}{\texttt{Qwen3-4B}} \\
Vanilla  & 32.27 & 23.99 & 60.50 & 45.50 & 87.40 & 75.50 & 78.70 & 57.69 \\
GRPO     & 36.15 & 28.77 & 61.30 & 43.90 & 87.10 & 75.00 & 79.30 & 58.79 \\
\rowcolor{table-blue!50}
\textbf{BuPO (Ours)} & \textbf{37.52} & \textbf{29.89} & \textbf{61.70} & \textbf{48.00} & \textbf{87.50} & \textbf{76.80} & \textbf{82.90} & \textbf{60.62} \\
\rowcolor{table-blue!50}
$\Delta$ vs.\ GRPO & \textcolor{blue}{+1.37} & \textcolor{blue}{+1.12} & \textcolor{blue}{+0.40} & \textcolor{blue}{+4.10} & \textcolor{blue}{+0.40} & \textcolor{blue}{+1.80} & \textcolor{blue}{+3.60} & \textcolor{blue}{+1.83} \\
\midrule
\multicolumn{9}{l}{\texttt{Qwen3-8B}} \\
Vanilla  & 32.51 & 22.45 & 63.40 & 51.00 & 90.80 & 79.60 & 83.54 & 60.47 \\
GRPO     & 40.31 & 31.55 & 66.60 & 52.00 & 90.87 & \textbf{81.55} & 84.15 & 63.86 \\
\rowcolor{table-blue!50}
\textbf{BuPO (Ours)} & \textbf{40.63} & \textbf{31.92} & \textbf{66.80} & \textbf{53.50} & \textbf{91.00} & 81.50 & \textbf{85.98} & \textbf{64.48} \\
\rowcolor{table-blue!50}
$\Delta$ vs.\ GRPO & \textcolor{blue}{+0.32} & \textcolor{blue}{+0.37} & \textcolor{blue}{+0.20} & \textcolor{blue}{+1.50} & \textcolor{blue}{+0.13} & -0.05 & \textcolor{blue}{+1.83} & \textcolor{blue}{+0.62} \\
\bottomrule
\end{tabular}
\caption{Out-of-domain evaluation results across general reasoning and coding benchmarks. \textbf{Bold} denotes the best.}
\label{tab:ood_eval}
\end{table*}

\section{Extended Experiment Results}
\subsection{Out-of-Domain Generalization}
\label{app:ood}
To evaluate the generalization ability of BuPO, we extend our evaluation to broader benchmarks spanning advanced math (OlympiadBench~\citep{he2024olympiadbench}, Minerva~\citep{lewkowycz2022solving}), general reasoning (MMLU-Pro~\citep{wang2024mmlu}, GPQA-D~\citep{rein2024gpqa}, ARC-C~\citep{clark2018think}, BBH~\citep{suzgun2023challenging}), and coding (HumanEval~\citep{chen2021evaluating}). As shown in Table~\ref{tab:ood_eval}, BuPO consistently outperforms both the vanilla baseline and GRPO across all benchmarks on both model scales. BuPO achieves average gains of +1.83 and +0.62 points over GRPO on Qwen3-4B and Qwen3-8B, respectively, with particularly notable improvements on GPQA-D (+4.10 for 4B, +1.50 for 8B) and HumanEval (+3.60 for 4B, +1.83 for 8B). These consistent gains across diverse tasks suggest that optimizing internal layer policies encourages the model to develop more transferable reasoning capabilities rather than overfitting to training task patterns.

\subsection{Comparison with Conventional Layer-aware Optimization}
To examine whether BuPO specifically benefits from internal policy optimization, rather than merely from emphasizing lower layers, we compare it with two conventional layer-aware GRPO variants. The first baseline freezes layers up to the selected layer $l$, while the second applies a smaller learning rate ($0.1\times$) to layers above $l$. As shown in Table~\ref{tab:additional_baselines}, BuPO consistently outperforms both alternatives on \texttt{Qwen3-4B} and \texttt{OctoThinker-3B}. These results show that simply freezing layers or reweighting learning rates is insufficient to match BuPO. In contrast, optimizing the internal policy through IPA provides a more effective training signal, confirming the effectiveness of BuPO.

\subsection{Multi-seed Robustness}
\label{app:multi_seed}

To address whether the reported improvements depend on a single random seed, we rerun both GRPO and BuPO with three random seeds under the same training setup. Table~\ref{tab:multi_seed} reports the mean and standard deviation across seeds. BuPO outperforms GRPO on most benchmarks. Averaged over the four benchmarks, BuPO improves over GRPO by +3.77, +2.53, +1.95, and +3.62 points on \texttt{Qwen3-4B}, \texttt{Qwen3-8B}, \texttt{OctoThinker-3B}, and \texttt{OctoThinker-8B}, respectively.

The standard deviations are generally small relative to the performance gaps, indicating that the improvements are stable across random seeds. These results show that the gains of BuPO are not an artifact of a single run and further support the reliability of the main results.

\begin{table}[t]
\centering
\resizebox{\columnwidth}{!}{
\begin{tabular}{lcccc}
\toprule
\textbf{Methods} & \textbf{AMC} & \textbf{MATH} & \textbf{AIME24} & \textbf{AIME25} \\
\midrule
\multicolumn{5}{l}{\texttt{Qwen3-4B}} \\
BuPO & \textbf{80.78$\pm$0.41} & \textbf{85.19$\pm$0.26} & \textbf{36.77$\pm$0.11} & \textbf{30.99$\pm$0.63} \\
GRPO & 76.67$\pm$0.36 & 83.12$\pm$0.62 & 32.15$\pm$0.47 & 26.70$\pm$1.93 \\
\midrule
\multicolumn{5}{l}{\texttt{Qwen3-8B}} \\
BuPO & \textbf{89.06$\pm$0.72} & 88.00$\pm$0.22 & \textbf{53.85$\pm$0.18} & \textbf{34.90$\pm$0.48} \\
GRPO & 86.31$\pm$0.39 & \textbf{88.02$\pm$0.06} & 48.19$\pm$1.36 & 33.19$\pm$0.30 \\
\midrule
\multicolumn{5}{l}{\texttt{OctoThinker-3B}} \\
BuPO & \textbf{27.71$\pm$0.24} & \textbf{49.88$\pm$0.17} & \textbf{0.66$\pm$0.06} & \textbf{3.86$\pm$0.08} \\
GRPO & 27.29$\pm$0.18 & 46.42$\pm$0.31 & 0.45$\pm$0.16 & 0.17$\pm$0.06 \\
\midrule
\multicolumn{5}{l}{\texttt{OctoThinker-8B}} \\
BuPO & \textbf{38.08$\pm$0.59} & \textbf{62.02$\pm$0.05} & \textbf{4.76$\pm$0.12} & \textbf{6.67$\pm$0.11} \\
GRPO & 35.16$\pm$0.41 & 57.12$\pm$0.21 & 2.64$\pm$0.24 & 2.15$\pm$0.06 \\
\bottomrule
\end{tabular}
}
\caption{Multi-seed results of BuPO and GRPO. We report mean $\pm$ standard deviation over three random seeds.}
\label{tab:multi_seed}
\end{table}

\begin{table}[t]
\centering
\resizebox{\columnwidth}{!}{
\begin{tabular}{lccccc}
\toprule
\textbf{Methods} & \textbf{AMC} & \textbf{MATH} & \textbf{AIME24} & \textbf{AIME25} & \textbf{Avg.} \\
\midrule
\multicolumn{6}{l}{\texttt{Qwen3-4B}} \\
BuPO & \textbf{81.09} & \textbf{84.90} & \textbf{36.88} & \textbf{31.15} & \textbf{58.51} \\
Frozen layers & 78.91 & 83.60 & 31.04 & 29.06 & 55.65 \\
Layerwise LR & 80.86 & 84.00 & \textbf{36.88} & 30.73 & 58.12 \\
\midrule
\multicolumn{6}{l}{\texttt{OctoThinker-3B}} \\
BuPO & \textbf{27.50} & \textbf{49.79} & \textbf{0.63} & \textbf{0.42} & \textbf{19.59} \\
Frozen layers & 23.67 & 46.40 & 0.31 & \textbf{0.42} & 17.70 \\
Layerwise LR & 26.17 & 44.80 & \textbf{0.63} & 0.21 & 17.95 \\
\bottomrule
\end{tabular}
}
\caption{Comparison with conventional layer-aware GRPO alternatives.}
\label{tab:additional_baselines}
\end{table}

\subsection{Internal Policy Entropy Dynamics for More Models}
In this section, we present additional preliminary analyses of internal policy entropy dynamics across a broader set of models. Specifically, we examine different variants of the same backbone, including Base, Instruct, and Mix versions, as well as models trained with supervised fine-tuning (SFT) and reinforcement learning (RL). In addition, we include the DeepSeek-Math model~\citep{he2025deepmath} to further enrich the comparative analysis.

\paragraph{\textbf{Further Training Has Limited Impact on Internal Reasoning Patterns.}}
We further analyze models that undergo additional training beyond standard pre-training. Specifically, we include \texttt{DeepSeek-R1-Distill-Qwen-7B}, which is further trained from \texttt{Qwen2.5-Math-7B} using distilled responses from~\citet{guo2025deepseek} with SFT, as well as \texttt{Llama-OctoThinker-3B-Base} and \texttt{Llama-OctoThinker-8B-Base}, which are obtained via continued pre-training (i.e., mid-training) based on \texttt{Llama-3.2-3B-Base} and \texttt{Llama-3.2-8B-Base}, respectively~\citep{wang2025octothinker}. 
Moreover, \texttt{Qwen3-4B} and \texttt{Qwen3-4B-Base} also show consistent pattern after post-training with RL. Notably, these additional training procedures exhibit only marginal influence on the internal reasoning patterns.

In summary, reinforcement learning, supervised fine-tuning, and mid-training do not substantially alter the model’s internal reasoning mechanisms, suggesting that these intrinsic patterns are primarily determined by the model architecture and initial pre-training. Understanding how such patterns emerge remains an important direction for future work.

\paragraph{\textbf{Same Series of Models Exhibit Consistent Structures.}} After analyzing in all analysis plots across models, we find that models from the same series show consistent structures. For instance, all Qwen3 series models show progressive internal reasoning patterns including \texttt{Qwen3-4B}, \texttt{Qwen3-8B} and \texttt{Qwen3-14B}, also with other base or instruct version. Also, Llama-3.1 and Llama-3.2 show intra-difference and inter-consistency. 

\begin{figure}[!t]
    \centering
    \includegraphics[width=0.48\columnwidth]{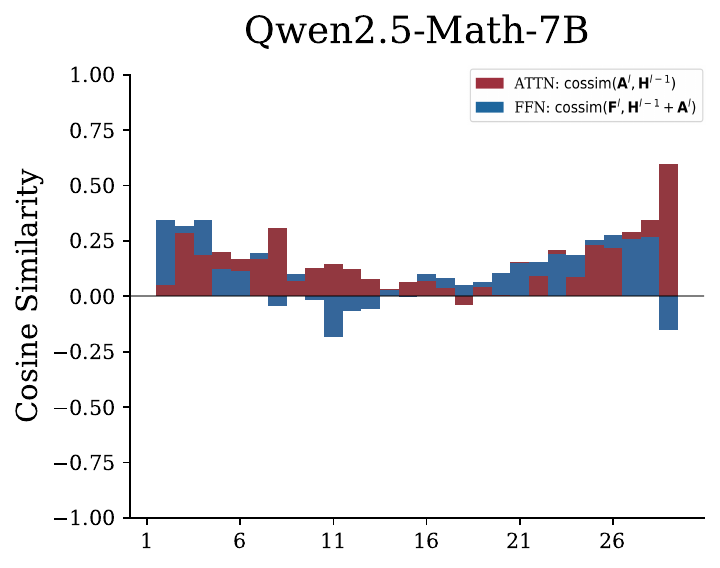}
    \hfill
    \includegraphics[width=0.48\columnwidth]{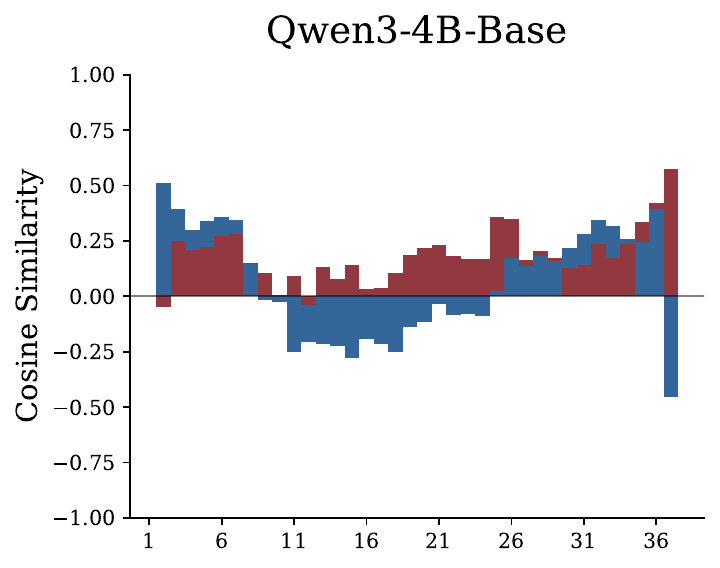}

    \vspace{0.4em}

    \includegraphics[width=0.48\columnwidth]{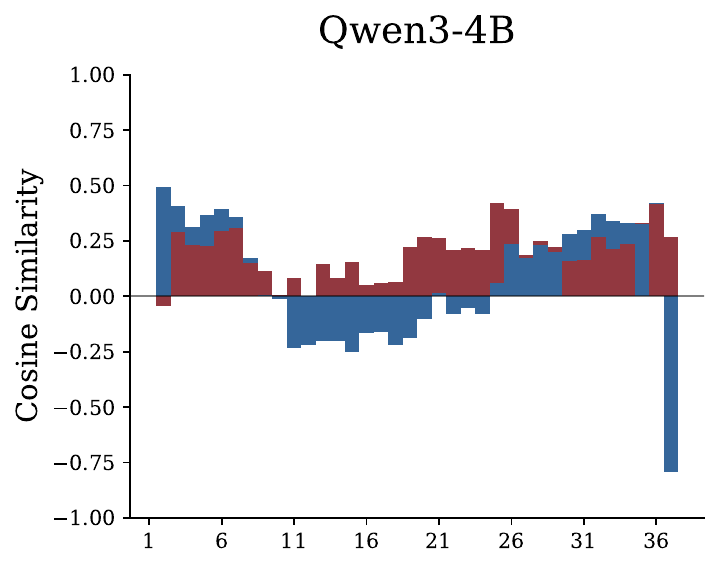}
    \hfill
    \includegraphics[width=0.48\columnwidth]{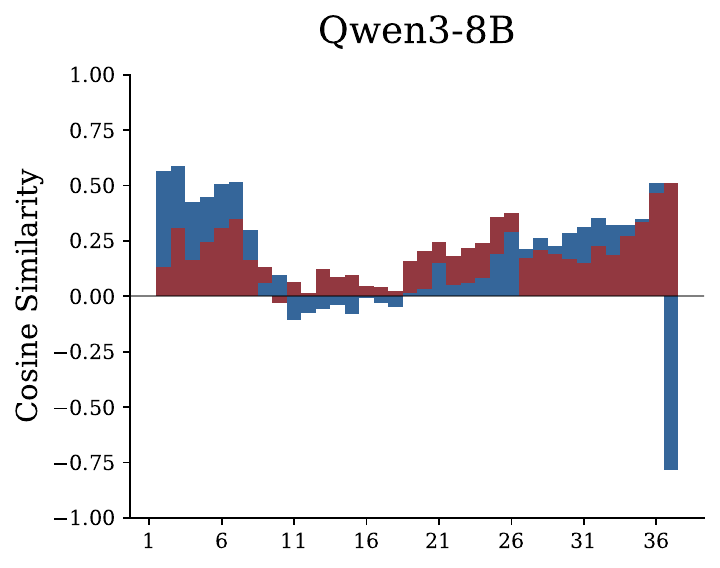}

    \caption{Residual cosine similarity across different Qwen models.}
    \label{fig:residual_similarity}
\end{figure}
\begin{figure*}[!h]
    \centering
    \subfigure{%
        \includegraphics[width=0.333\textwidth]{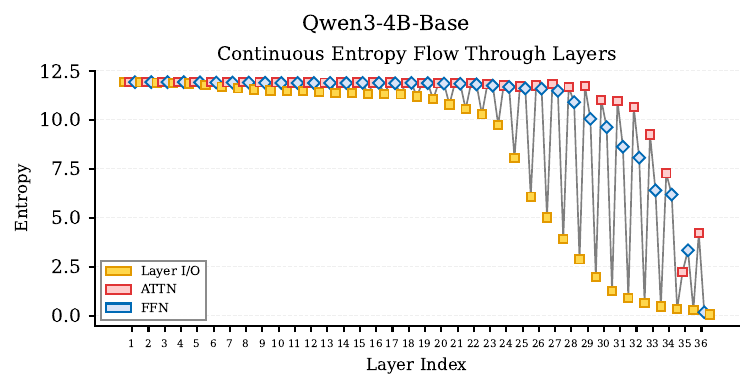}
    }%
    \subfigure{%
        \includegraphics[width=0.333\textwidth]{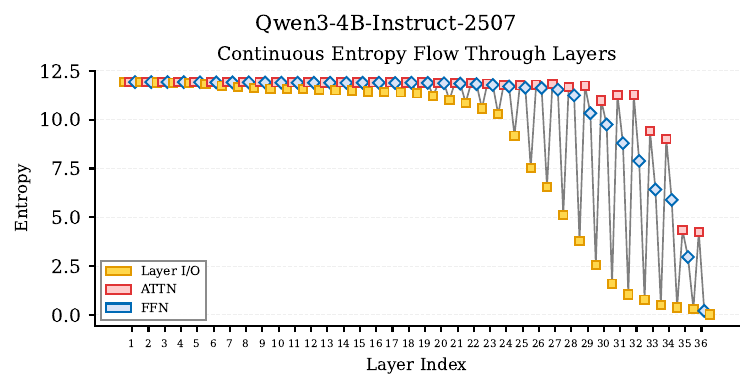}
    }%
    \subfigure{%
        \includegraphics[width=0.333\textwidth]{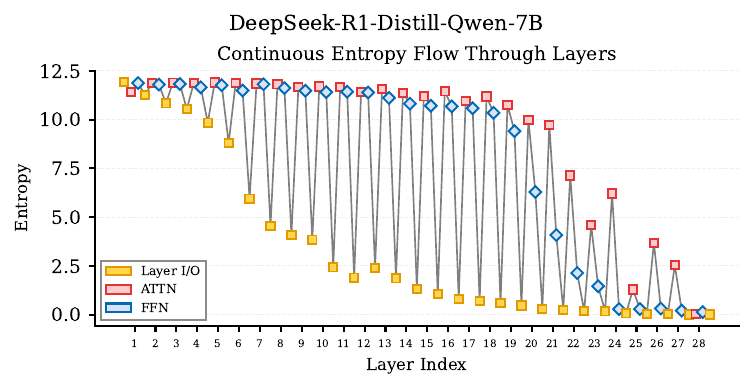}
    }%
    
    \subfigure{%
        \includegraphics[width=0.333\textwidth]{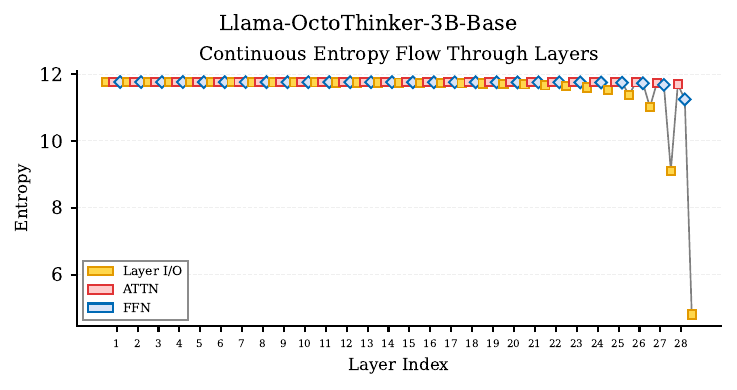}
    }%
    \subfigure{%
        \includegraphics[width=0.333\textwidth]{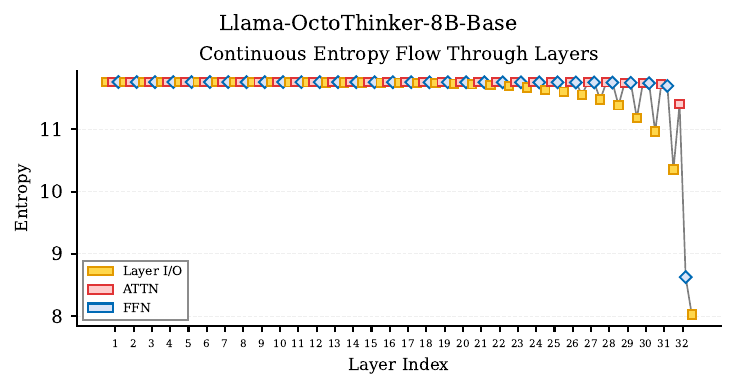}
    }%
     \subfigure{%
        \includegraphics[width=0.333\textwidth]{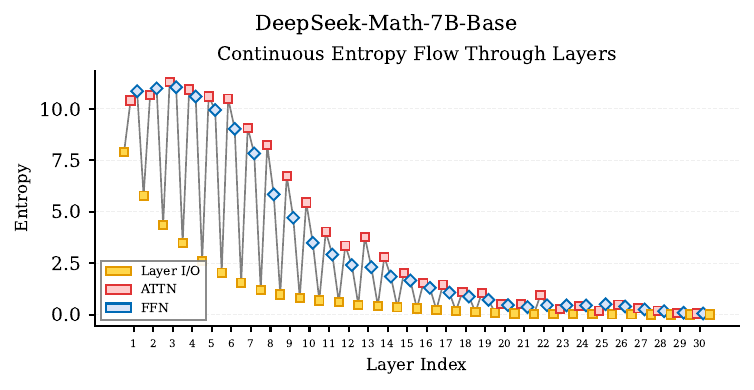}
    }%
    \caption{Continuous entropy dynamics of internal policy for additional models. The information flows from $\mathbf{H}^{l-1}$ into $\mathbf{A}^l$, $\mathbf{F}^l$, then to the next layer $\mathbf{H}^l$.}
    \label{fig:interal_entropy_flow_app}
\end{figure*}
\begin{figure*}[t!]
    \centering
     \subfigure{%
        \includegraphics[width=\textwidth]{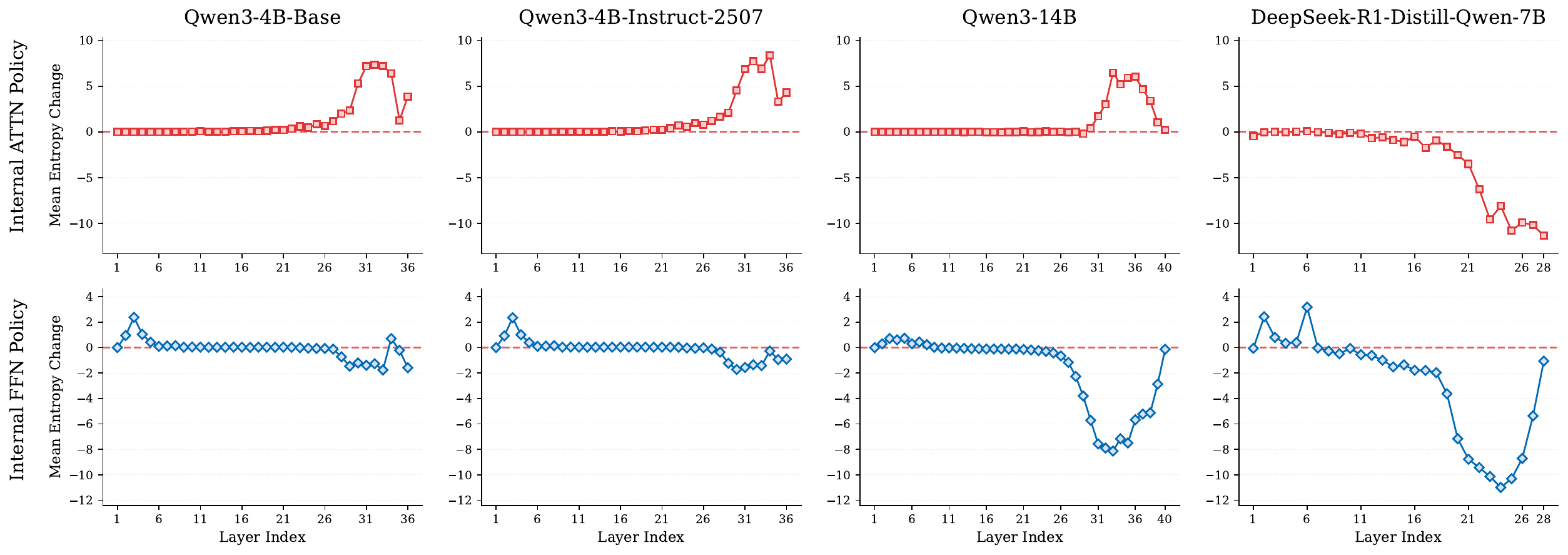}
    }
    \subfigure{%
        \includegraphics[width=\textwidth]{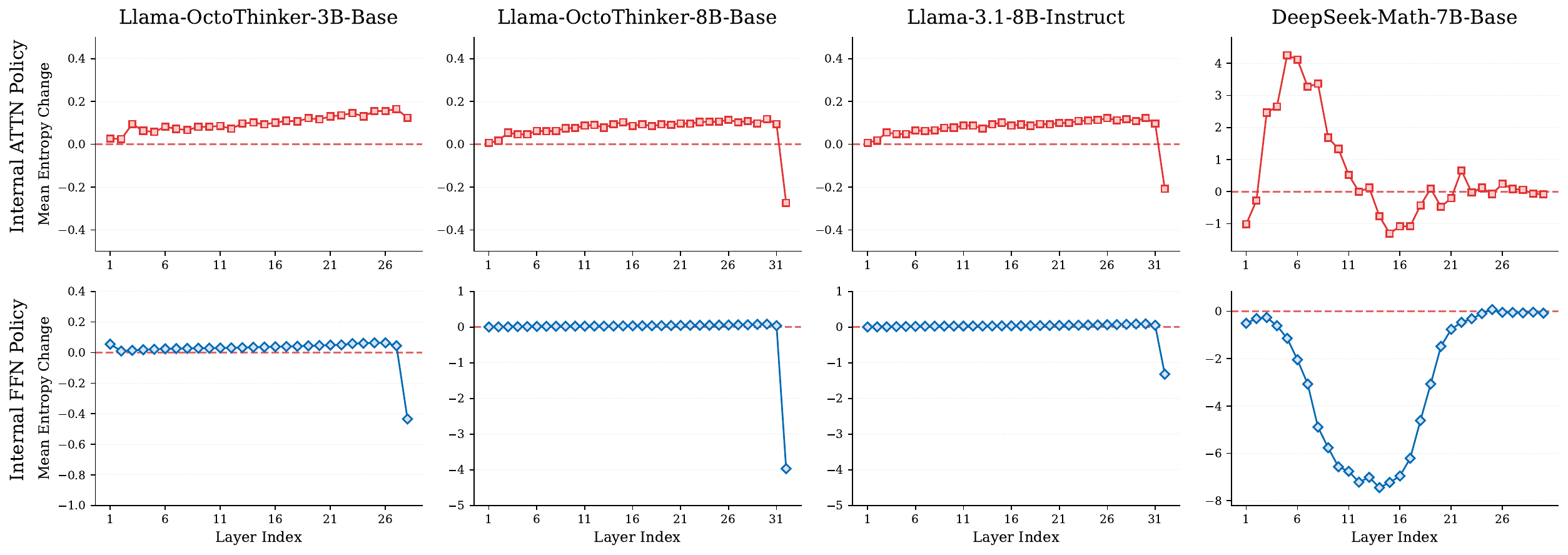}
    }
    \caption{Entropy change dynamics of internal policy with more models. }
    \label{fig:modular_entropy_app1}
\end{figure*}

\paragraph{\textbf{Entropy Dynamics of DeepSeek-Math.}}
We further analyze the internal reasoning mechanisms of \texttt{DeepSeek-Math-7B-Base} to provide a more comprehensive comparison. As illustrated in Figure~\ref{fig:interal_entropy_flow_app}, \texttt{DeepSeek-Math-7B-Base} exhibits a markedly different entropy dynamics: the overall internal policy entropy decreases substantially and converges primarily in the middle layers. This phenomenon is primarily driven by the consistently negative entropy change in the FFN module, i.e., $\Delta H_{\text{FFN}}^l < 0$, as illustrated in Figure~\ref{fig:modular_entropy_app1}, particularly in the middle layers.

Based on this observation, we infer that both Qwen and DeepSeek-Math demonstrate strong capability in knowledge absorption during post-training, indicating that convergence behavior in internal reasoning plays a critical role in effective learning, in contrast to Llama. Moreover, the generation search space of DeepSeek-Math appears more constrained than that of Qwen, particularly the Qwen3 series, suggesting reduced exploration capacity. We hypothesize that such internal reasoning patterns significantly influence the effectiveness of further training, pointing to promising directions for architectural design and optimization of foundation models.

\subsection{How do Internal modules influence the residual stream?}
\label{app:internal_similarity}
To further understand how internal modules shape the residual stream in Qwen models with a progressive reasoning pattern, we analyze residual cosine similarity, which quantifies how each module writes to the residual pathway~\citep{hu2025affects}. For a given layer $l$, we compute $\text{cossim}(\mathbf{A}^l, \mathbf{H}^{l-1})$ for self-attention and $\text{cossim}(\mathbf{F}^l, \mathbf{H}^{l-1}+\mathbf{A}^l)$ for the FFN. A cosine similarity near zero indicates writing new, orthogonal features; negative values indicate feature suppression; and positive values indicate amplification of existing features.

As shown in Figure~\ref{fig:residual_similarity}, the Qwen models largely follow the entropy dynamics discussed earlier, while exhibiting clear inter-generation differences. For Qwen3, self-attention consistently amplifies the residual stream, in line with its positive entropy change and expanded exploration behavior. In contrast, Qwen2.5 shows noticeably weaker attention write-in strength, with reduced cosine similarity magnitudes, consistent with its negative entropy change in self-attention.

The FFN modulates the residual stream in a stage-dependent manner across all models: in lower layers, it injects largely orthogonal features to support exploration; in middle layers, it suppresses vague signals while integrating parametric knowledge in FFN, corresponding to the Integration stage; and in upper layers, it amplifies and integrates features to drive convergence. Across all models, the final layer exhibits a sharp directional shift, underscoring its critical role in final prediction~\citep{gupta2025llms, agarwal2025unreasonable}.

\newpage
\subsection{Pass@$K$ Performance across Datasets}
We further provide a detailed version of Figure~\ref{fig:passk_main} across AMC23, MATH500, AIME24, and AIME25. Our proposed BuPO consistently outperforms the vanilla GRPO baseline.

\begin{figure*}[h!]
    \centering
    \includegraphics[width=1\textwidth]{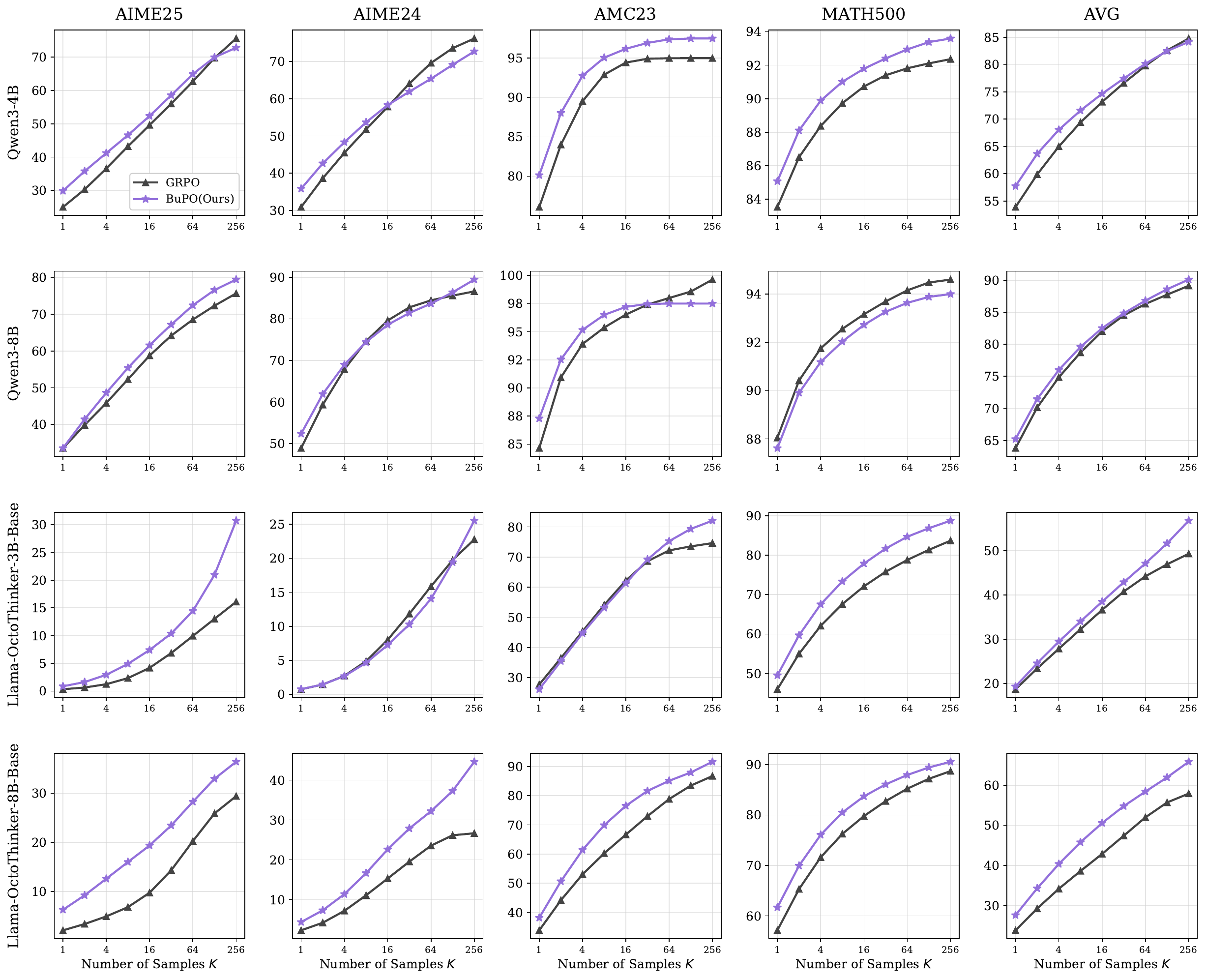}
    \caption{Pass@K results on MATH500, AMC23, AIME24 and AIME25. To reduce evaluation variance, we set $n=300$.}
    \label{fig:passk_app}
\end{figure*}
\section{Use of AI Assistants}

AI assistants were used only for minor language polishing and grammar refinement. All research ideas, technical contributions, experiments, analyses, and final claims were developed and verified by the authors.

\end{document}